%% file: COMBINER.tex
\documentclass[10pt,journal]{IEEEtran}
\usepackage{amsmath,amsfonts}
\usepackage{algorithmic}
\usepackage{array}
\usepackage{soul}
\usepackage{threeparttable}
\usepackage{color}
\usepackage{xcolor}
\usepackage{tikz}
\usepackage{pgfplots}
\usepackage{float}
\usepackage{textcomp}
\usepackage{stfloats}
\usepackage{url}
\usepackage{ulem}
\usepackage{verbatim}
\usepackage{graphicx}
\usepackage{subcaption}
\usepackage{cite}
\usepackage{CJK}
\usepackage[breaklinks,colorlinks,linkcolor=blue,citecolor=blue]{hyperref}

\usepackage[linesnumbered,algoruled,vlined]{algorithm2e}
\usepackage{booktabs}
\usepackage{multirow}
\usepackage{makecell}
\usepackage{longtable}
\usepackage{multicol}
\usepackage{utfsym}
\usepackage{hhline}
\usepackage{colortbl}
\usepackage{bm}
\usepackage{url}
\usepackage{enumitem}
\usepackage{arydshln}
\usepackage{bbding}
\usepackage{utfsym}
\usepackage{orcidlink}
\usepackage[switch]{lineno}

\definecolor{lightgray}{rgb}{0.88, 0.92, 0.98}
\definecolor{defblue}{rgb}{0.1843, 0.3333, 0.6}
\definecolor{defred}{rgb}{0.88, 0.2510, 0.3294}
\definecolor{green1}{rgb}{ 0.910,  0.953,  0.855}
\definecolor{green2}{rgb}{0.82,  0.902,  0.710}
\definecolor{green3}{rgb}{0.713,  0.903,  0.648}
\definecolor{green4}{rgb}{ 0.81,  0.933,  0.775}
\definecolor{green5}{rgb}{ 0.71,  0.913,  0.775}
\definecolor{green6}{rgb}{ 0.725,  0.855,  0.561}

\definecolor{defyellow}{rgb}{1,  0.983,  0.717}
\definecolor{defyellowtext}{rgb}{1,  0.851,  0.438}

\definecolor{revisecolor}{rgb}{0,  0,  0}
\definecolor{revisecolor2}{rgb}{0,  0,  0}
\definecolor{revisecolor3}{rgb}{0,  0,  0}

\usepackage{amsthm}
\theoremstyle{plain}

\usepackage{mathtools}

\DeclarePairedDelimiterX{\infdivx}[2]{(}{)}{%
  #1\;\delimsize\|\;#2%
}

\pgfplotsset{compat=1.18} 
\usepackage{environ}
\NewEnviron{myequation}{
    \begin{equation*}
    \scalebox{1}{$\BODY$}
    \end{equation*}
    }

\begin{document}
\begin{CJK}{UTF8}{gbsn}
\title{COMBINER: Composed Image Retrieval Guided by Attribute-based Neighbor Relations}


\author{Zixu~Li~\orcidlink{0009-0001-5136-159X},
        Yupeng~Hu~\orcidlink{0000-0002-5653-8286}\IEEEmembership{, Member, IEEE}, 
        Zhiwei~Chen~\orcidlink{0009-0003-0365-8553},
        Haokun~Wen~\orcidlink{0000-0003-0633-3722},\\
        Xuemeng~Song~\orcidlink{0000-0002-5274-4197}\IEEEmembership{, Senior Member, IEEE},
        Liqiang~Nie~\orcidlink{0000-0003-1476-0273}\IEEEmembership{, Senior Member, IEEE}

\thanks{Manuscript received 10 May 2025; revised 15 November 2025, 11 February 2026, and 26 March 2026; accepted 29 April 2026. This work was supported in part by the National Natural Science Foundation of China, No.:62576195, No.:62276155, No.:62376137 and No.:624B2047; in part by the Key R\&D Program of Shandong Province (Major scientific and technological innovation projects), China, No.: 2025CXGC020101. \textit{(Corresponding author: Yupeng Hu.)}}
\thanks{\noindent Zixu Li,~Yupeng Hu,~Zhiwei Chen are with the School of Software, Shandong University, Jinan, 250100, China. (Email: \{lizixu.cs, zivczw\}@gmail.com; huyupeng@sdu.edu.cn)}
\thanks{Haokun Wen is with the School of Computer Science and Technology, Harbin Institute of Technology (Shenzhen), Shenzhen, 518000, China, and Department of Data Science, City University of Hong Kong, Hong Kong 523808, China (e-mail: whenhaokun@gmail.com). }
\thanks{Xuemeng Song is with the Department of Computer Science and Engineering, Southern University of Science and Technology, Shenzhen, 518000, China (e-mail: sxmustc@gmail.com).}
\thanks{Liqiang Nie is with the School of Computer Science and Technology, Harbin Institute of Technology (Shenzhen), Shenzhen, 518000, China. (Email: nieliqiang@gmail.com).}
}

\markboth{}%
{Li \MakeLowercase{\textit{et al.}}: COMBINER: Composed Image Retrieval Guided by Attribute-based Neighbor Relations}


\maketitle

\begin{abstract}
Composed Image Retrieval (CIR) represents a challenging retrieval task that targets locating specific images through multimodal inputs. Despite recent progress in CIR techniques, prior approaches often overlook cases where images appear visually alike yet differ in attributes, potentially undermining both multimodal feature fusion and similarity modeling. To mitigate this limitation, we design a unified representation of cross-modal features based on attribute prototypes. Nevertheless, the task is far from straightforward, owing to three core issues: (1) entanglement in attribute-level semantics, (2) inconsistency across modalities, and (3) supervised signal missing. To tackle the above obstacles, we introduce a COMposed image retrieval network guided By attrIbute-based NEighbor Relations (COMBINER). Specifically, we first design an Adaptive Semantic Disentanglement module, which is capable of disentangling attribute features based on multimodal primitive features. Secondly, we propose a Unified Prototype-based Composition module, which can construct cross-modal unified prototypes (CUP) and facilitate multimodal feature composition. Finally, we introduce a Dual Relations Modeling module, which can mine pairwise and neighbor relations based on attribute similarity. Compared to traditional neighbor relations modeling CIR methods, COMBINER represents the first study addressing the phenomenon of visually similar but attribute-unrelated samples. It achieves a more accurate understanding of the semantic relations among samples by employing an attribute prototype-based similarity metric. Comprehensive experiments conducted on three benchmark datasets confirm the effectiveness of our proposed COMBINER. The implementation of our method will be accessed at https://github.com/Lee-zixu/COMBINER
\end{abstract}

\begin{IEEEkeywords}
Composed image retrieval, Multimodal fusion, Multimodal retrieval
\end{IEEEkeywords}

\input{1_int.tex}

\input{2_rel.tex}

\input{3_met.tex}

\input{4_exp.tex}

\input{5_con.tex}

\normalem
\bibliographystyle{ieee_fullname}
\bibliography{reference}

\end{CJK}
\end{document}

%% file: 1_int.tex
\section{Introduction}
\IEEEPARstart{A}{s} the number of images grows exponentially~\cite{TempRet,STABLE,ERASE}, the challenge of searching for images that match user interests has increasingly attracted attention in the realm of information retrieval~\cite{image-text-TIP-1, image-text-TIP-2, image-text-TIP-3, image-text-TIP-4, REFINE}. Nevertheless, traditional image retrieval methods are constrained by limited query conditions, such as using only pure image or text-based queries, making them incapable of meeting the broad needs of users. In response to this issue, Vo \textit{et al.}~\cite{tirg} introduced the concept of Composed Image Retrieval (CIR), which empowers users to locate the desired image through a more adaptable composed query that integrates both a reference image and a modification description~\cite{HINT,MELT,Air-Know,ConeSep}. The reference image in this setup is a rough approximation of the user's intended image, but does not perfectly match the target, whereas the modification description specifies the variations between the reference and target images as envisioned by the user. CIR's effective expression of retrieval intent makes it highly relevant for intelligent retrieval tasks, including semantic understanding~\cite{shi2026mmerror,EgoAction,dai2025detecting, li2024mlp, tang2026mwd,zhang2025event,EgoAdapt,OmniEgo-R,R3,tian2025core} and multimodal learning~\cite{lu2026chordedit,hu2023semantic,zhao2026information,hu2021coarse,ruan2026grain, zhang2026angel,hu2021video}.

\begin{figure}[t]
	\includegraphics[width=0.98\linewidth]{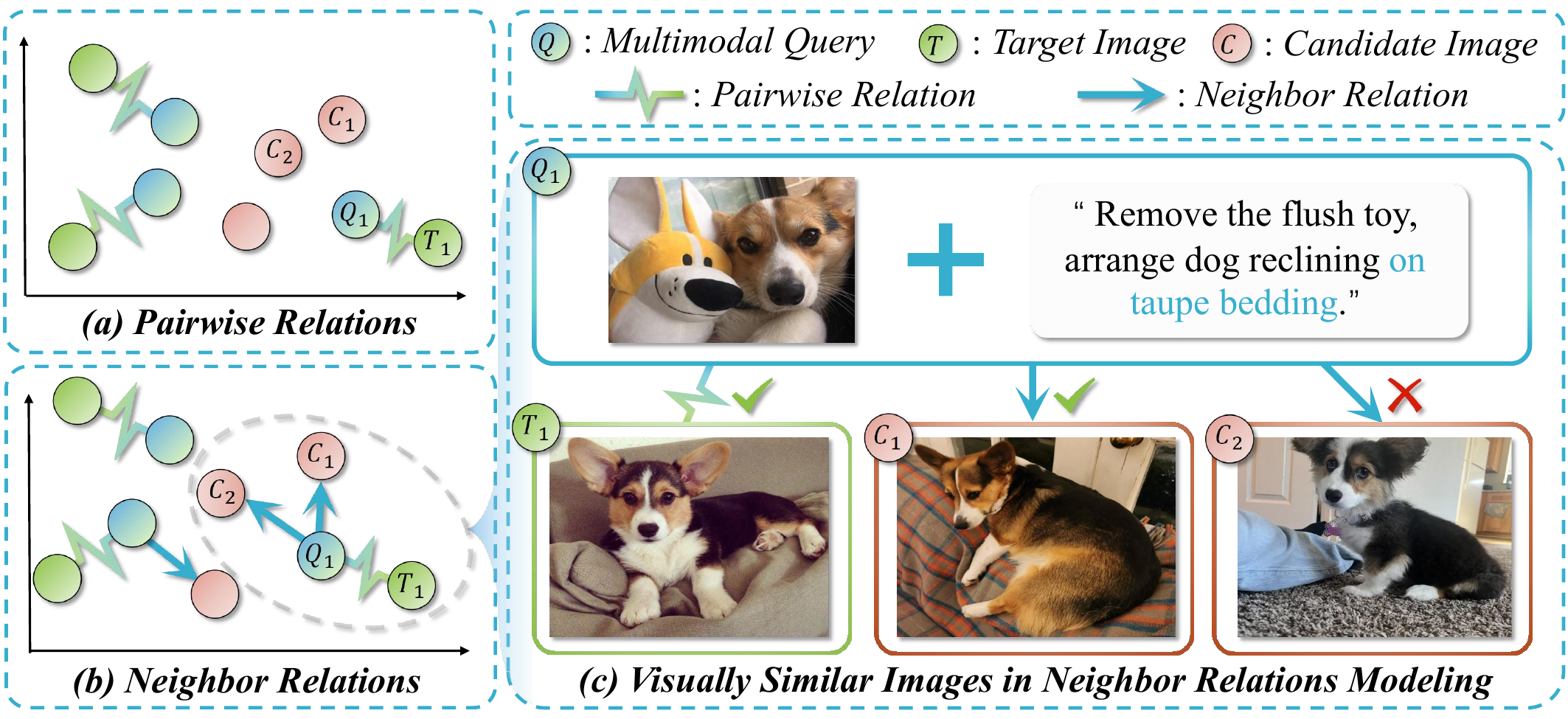}
\vspace{-6pt}
	\caption{Example of (a) Pairwise Relations, (b) Neighbor Relations, and (c) Visually Similar Images in Relations Modeling. In this figure, $\textit{\textbf{Q}}$ denotes the multimodal query, $\textit{\textbf{T}}$ denotes the target image, and $\textit{\textbf{C}}$ denotes the candidate image. Fig.~1(c) illustrates the traditional neighbor relations modeling methodology brings both candidate images $\textit{\textbf{C}}_{\textbf{1}}$ and $\textit{\textbf{C}}_{\textbf{2}}$ close to $\textit{\textbf{Q}}_{\textbf{1}}$. However, $\textit{\textbf{C}}_{\textbf{2}}$ is visually similar but attribute-unrelated with $\textit{\textbf{Q}}_{\textbf{1}}$ (``carpet'' does not match the query ``bedding''). Therefore, $\textit{\textbf{C}}_{\textbf{2}}$ should not be brought close to $\textit{\textbf{Q}}_{\textbf{1}}$.}
\vspace{-20pt}
	\label{fig:intro}
\end{figure}

\textcolor{revisecolor}{In recent years, although significant progress has been made in CIR research, existing studies often overlook the importance of distinguishing visually similar but attribute-irrelevant images. Specifically, some studies~\cite{Prog-Lrn, sprc, ssn} tend to focus only on the pairwise relationship between the multimodal query and the target image.}
 As illustrated in Fig.~\ref{fig:intro}(a), the pairwise relation merely brings the multimodal query close to its corresponding unique target image while pushing all other candidate images away as negative samples. Due to visual similarity among images, a significant portion of candidate images can be classified as false-negative samples, yet they are erroneously pushed away during the pairwise relations learning process~\cite{false-negative1, false-negative2}. To mitigate the impact of false-negative samples on metric learning, some studies~\cite{tgcir, SADN} have explored neighbor relations. 
 
 As shown in Fig.~\ref{fig:intro}(b), modeling neighbor relations can reduce the distance between multimodal queries and visually similar candidate images. However, not all visually similar images are false-negative samples. For instance, in Fig.~\ref{fig:intro}(c), both images $\textit{\textbf{C}}_{\textbf{1}}$ and $\textit{\textbf{C}}_{\textbf{2}}$ exhibit high visual similarity with the target image $\textit{\textbf{T}}_{\textbf{1}}$, however, the $\textit{\textbf{C}}_{\textbf{2}}$ image contains a ``carpet'' instead of the ``bedding'' required in the modification text. We refer to images like $\textit{\textbf{C}}_{\textbf{2}}$ as visually similar but attribute-unrelated images. These images do not constitute false-negative samples and should be pushed away during the metric learning. 

\textcolor{revisecolor}{However, existing neighbor relationship modeling methods~\cite{tgcir, SADN} primarily rely on visual similarity, which leads to the model being unable to distinguish between false-negatives and visually similar but attribute-unrelated samples, erroneously pushing them closer in the space to visually similar multimodal queries, ultimately resulting in a decline in retrieval accuracy. For instance, SADN~\cite{SADN}, a representative work in this field, utilizes neighborhood-based knowledge distillation to significantly improve the inter-modal distribution differences. However, it relies on initial similarity to select neighbors. If the initial retrieval model is misled by visually similar but attribute-irrelevant distractors, SADN might aggregate these erroneous distractors as neighbors, thereby introducing noise. Therefore, these methods struggle to directly address the visually similar but attribute-irrelevant problem.}

\begin{figure}[t]
	\includegraphics[width=0.97\linewidth]{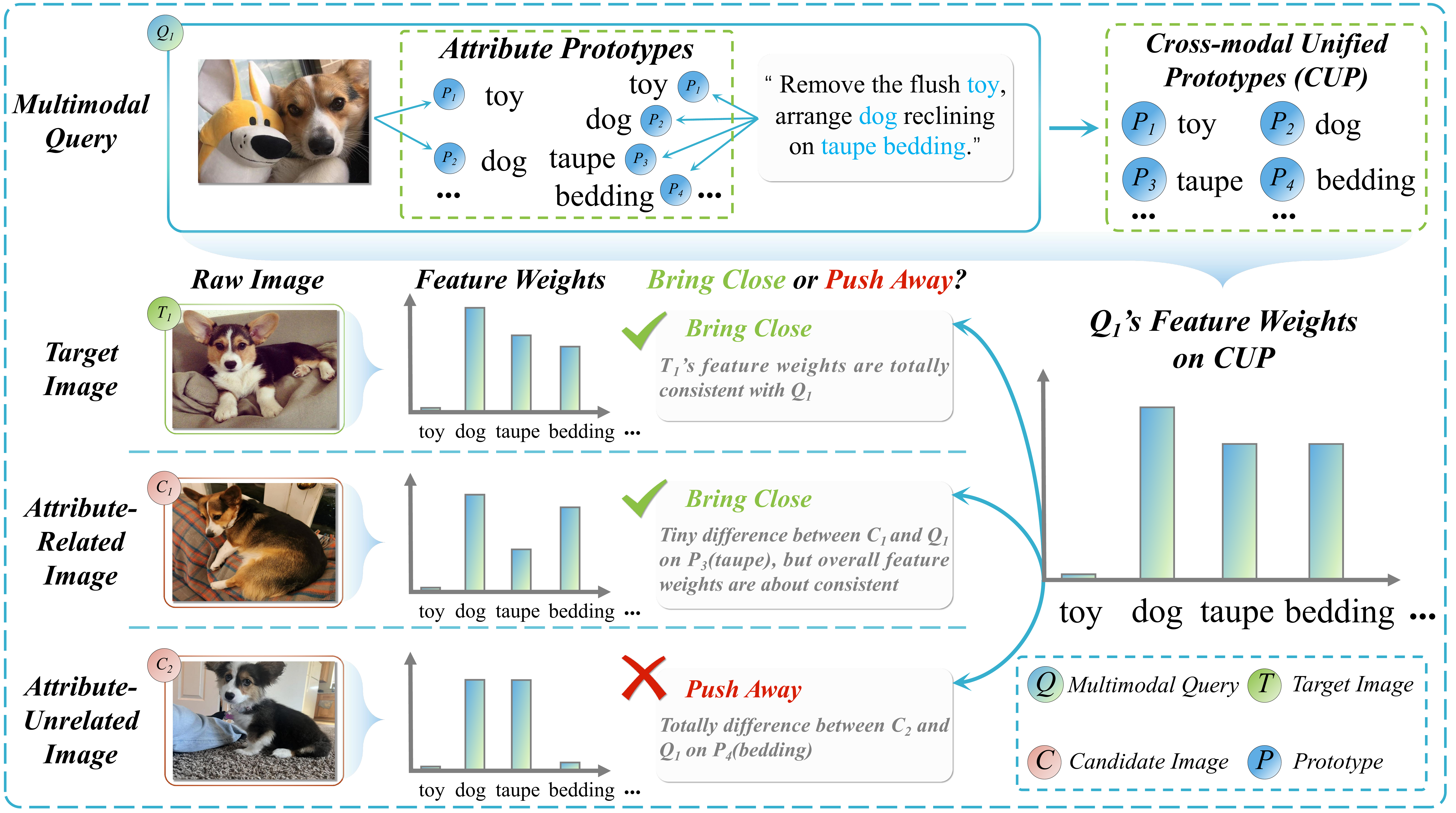}
\vspace{-4pt}
	\caption{Schematic of our proposed similarity measure method based on attribute prototypes.}
\vspace{-23pt}
	\label{fig:intro_2}
\end{figure}

To address the aforementioned problem, it is natural to think of incorporating sample attribute information into the similarity measure. 
\textcolor{revisecolor}{Existing metric learning methods that process attributes via prototypes or subspaces often employ static prototypes (for example, ADDE-M~\cite{R-5}), or they rely on predefined attribute labels to learn fixed editing directions or templates (for example, Face-StyleGAN~\cite{R-6} and AMGAN~\cite{R-7}). In other words, these methods depend on fine-grained, explicit attribute annotations to provide strong supervision signals. This confines their application to closed-domain datasets possessing fixed attribute labels, making them difficult to apply in open-domain CIR scenarios that lack attribute annotations. Furthermore, the aforementioned methods typically assume that visual similarity within non-target subspaces, such as shape or identity information, should always be preserved. This assumption often proves ineffective when confronted with visually similar but attribute-irrelevant distractors. To eliminate the influence of visually similar but attribute-irrelevant samples on CIR models and simultaneously avoid the limitations of the aforementioned methods,}
inspired by the proverb ``\textit{Like attracts like}'', we introduce a unified representation of cross-modal features based on attribute prototypes, which is beneficial to identify visually similar and attribute-related samples while pushing away attribute-unrelated samples to optimize multimodal feature fusion and metric learning. 

As illustrated in Fig.~\ref{fig:intro_2}, our approach first extracts sets of attribute prototypes from the image and text modalities, respectively. Then, through cross-modal semantic interaction, the union of the two prototype sets is obtained (\textit{e.g.}, toy, dog, taupe, bedding), which we refer to as the \uline{C}ross-modal \uline{U}nified \uline{P}rototype (CUP). Subsequently, we compute the distribution weights of each sample's features on the CUP to assess attribute similarity among different samples, which is then utilized in the metric learning process. Based on the differences in attribute similarity, we classify all visually similar samples into three categories: 1) \textbf{Target images like $\textit{\textbf{T}}_{\textbf{1}}$}. Intuitively, the target image $\textit{\textbf{T}}_{\textbf{1}}$'s feature weights are most similar to those of $\textit{\textbf{Q}}_{\textbf{1}}$, and they should be brought close in the metric space. 2) \textbf{Attribute-related images like $\textit{\textbf{C}}_{\textbf{1}}$}. Although there are some differences in the weights of the ``taupe'' attribute between $\textit{\textbf{C}}_{\textbf{1}}$ and $\textit{\textbf{Q}}_{\textbf{1}}$, the overall feature weights are relatively close, so they should be brought close together. 3) \textbf{Attribute-unrelated images like $\textit{\textbf{C}}_{\textbf{2}}$}. Despite this type of images show visual similarity to $\textit{\textbf{Q}}_{\textbf{1}}$, the weights for the ``bedding'' attribute are significantly different,therefore, $\textit{\textbf{C}}_{\textbf{2}}$ is classified as a visually similar but attribute-unrelated image in relation to $\textit{\textbf{Q}}_{\textbf{1}}$ and should be pushed away.

However, implementing the aforementioned approach is non-trivial because of three primary obstacles. 1) \textbf{Entanglement in attribute-level semantics}. Whether dealing with the image or text, the semantic information of different attributes is often entanglement and does not naturally exist in independent forms. Consequently, the primary challenge lies in disentangling the various attribute prototypes from the original features. 2) \textbf{Inconsistency across modalities}. Since the image and text belong to distinct modalities, their attribute prototypes may be inconsistent, creating difficulties in multimodal query feature fusion and metric learning. Therefore, the second challenge is how to interact the information from different modal and extract the CUP. 3) \textbf{Supervised signal missing}. To identify visually similar but attribute-unrelated images, we model pairwise and neighbor relations based on attribute similarity. However, we lack explicit supervisory signals for neighbor relations. Therefore, the third challenge is how to model neighbor relations based on attribute similarity in the absence of supervisory signals.

To tackle the aforementioned difficulties, we introduce an innovative \textbf{COM}posed image retrieval network guided \textbf{B}y attr\textbf{I}bute-based \textbf{NE}ighbor \textbf{R}elations \mbox{(COMBINER)}, which facilitates multimodal feature fusion and metric learning through an attribute prototype-based similarity metric. Specifically, we first design an Adaptive Semantic Disentanglement module to isolate attribute features. Second, we introduce a Unified Prototype-based Composition module to construct the CUP and derive attribute-level composed features. Finally, we develop a Dual Relations Modeling module to model pairwise and neighbor relations based on attribute similarity, thereby optimizing the metric learning process.

The principal contributions of this paper are as follows:
\begin{itemize}[leftmargin=8pt]
    \item To the best of our understanding, we are the first to explore the phenomenon of visually similar but attribute-unrelated images that exists in the CIR task. We also propose a unified representation of cross-modal features based on attribute prototypes, which is beneficial to mitigate the negative impact of this phenomenon on metric learning.
    \item We design a novel CIR model, COMBINER, to achieve effective multimodal query feature fusion and metric learning based on attribute prototypes through adaptive semantic disentanglement, unified prototype-based composition, and dual relations modeling.
    \item Extensive experiments on three CIR benchmark datasets demonstrate that COMBINER achieves optimal performance across all metrics, proving its superiority and the effectiveness of its components. Before the introduction of COMBINER, the highest results on each benchmark were obtained by different approaches, such as DQU-CIR on Shoes and FashionIQ datasets, and SPRC on CIRR dataset.
\end{itemize}

%% file: 2_rel.tex
\begin{table*}[ht!]
  \centering\small
      \renewcommand\arraystretch{1.2}
  \caption{\textcolor{revisecolor}{Symbol Table of our proposed COMBINER.}}
  \vspace{-4pt}
  \tabcolsep=2pt
  \label{tab:Symbol_Table}
  \resizebox{0.85\linewidth}{!}{
  \begin{tabular}{c|l|c||c|l|c}
    \hline
    \hline
    \textbf{Symbol} & \textbf{Explanation} & \textbf{Ref.} & 
    \textbf{Symbol} & \textbf{Explanation} & \textbf{Ref.}  
     \\
    \hline
    $\mathcal{T}$ & Set of $N$ training triplets & Sec.~\ref{sec:pre} &
    $\textbf{W}_c$ & Prototype weights learned by SAA & Eq.~\ref{eq:eq3}  \\
    
    $x_r$ & Reference Image & Sec.~\ref{sec:pre} &
     $\hat{\textbf{w}}_r$ & Image prototype weights computed by UPC & Eq.~\ref{eq:eq6}  \\
    
    $t_m$ & Modification Text & Sec.~\ref{sec:pre} &
     $\hat{\textbf{w}}_m$ & Text prototype weights computed by UPC & Eq.~\ref{eq:eq6}   \\
    
    $x_t$ & Target Image & Sec.~\ref{sec:pre} &
     $\textbf{F}_c$ & Final attribute-level composed features & Eq.~\ref{eq:eq7}    \\
    
    $D$ & Feature embedding dimension & Sec.~\ref{sec:asd}  & 
    $\bar{\textbf{F}}_{c}$ & Average-pooled composed features & Eq.~\ref{eq:eq8}   \\
    
    $X$ & Number of global attribute prototypes & Sec.~\ref{sec:asd}  &
    $\bar{\textbf{F}}_{t}$ & Average-pooled target features & Eq.~\ref{eq:eq8}   \\
    
    $Y$ & Number of local attribute prototypes & Sec.~\ref{sec:asd}  &
    $H$ & Number of semantic cluster centers & Sec.~\ref{sec:udrm}  \\
    
    $U$ & Total number of attribute prototypes ($U=X+Y$) & Sec.~\ref{sec:asd}  &
    $\mathcal{P}$ & Set of semantic cluster centers & Sec.~\ref{sec:udrm}  \\

    \hline
    \hline
  \end{tabular}
}
    \vspace{-21pt}
\end{table*}

\section{Related Work}

\subsection{Composed Image Retrieval}
As introduced earlier, Composed Image Retrieval is designed to locate the target image that aligns with users’ intent via the multimodal query. This paradigm facilitates a more precise alignment with complex user demands and has drawn considerable attention in recent literature~\cite{crn,eer,afce,FineCIR,pair,median,sprc,candidate,dqu,lincir,seize,magiclens,OFFSET,HUD}.  
Based on the backbone architectures adopted for feature encoding, existing CIR approaches can generally be categorized into two main types: those relying on conventional models and those built upon visual-language pretraining (VLP) techniques.  
The first category~\cite{tirg, mgur, cosmo, clvcnet, crn} typically applies neural architectures such as ResNet or LSTM to project visual and textual inputs into distinct embedding spaces. Subsequently, they employ cross-modal composition modules to integrate the semantics from both image and text representations.
For instance, Lee \textit{et al.}~\cite{cosmo} proposed a method via adjusting the reference image locally to meet content modification demands, followed by injecting global style attributes to fulfill stylistic transformation goals. 
The second category of approaches~\cite{limn, tgcir, TEMA, sprc, candidate, HABIT, SDQUR, INTENT, ReTrack} employs VLP-based models, such as CLIP~\cite{clip}, as the backbone for feature encoding. These models have seen extensive application in CIR research in recent years, delivering strong performance attributed to their robust feature encoding capabilities.
For example, CLIP has been employed~\cite{lfclip} to obtain visual and textual representations, which are then composed using basic linear transformations and ReLU activations to deliver strong performance in retrieval tasks. 
\textcolor{revisecolor2}{While these approaches have led to notable advancements, they predominantly rely on global feature alignment. A critical limitation is their inability to distinguish hard negatives, which refer to candidates that share high visual similarity with the reference image but contradict the modification text (e.g., wrong color or texture). Methods like SADN~\cite{SADN} attempt to use neighbor information but still depend on global visual similarity, which often leads to the aggregation of these attribute-unrelated distractors as false positives.}
In contrast, our proposed COMBINER framework exploits attribute prototypes to support dual relational modeling, effectively distinguishing between neighbors that are visually similar but attribute-unrelated.

\subsection{Prototype Learning}
Prototype learning focuses on acquiring prototype vectors that serve to express core prototype content, such as class labels and attribute semantics. This paradigm has attracted increasing attention in the deep learning community~\cite{zhu2026high,liu2025trainingfreemultistylefusionreferencebased,li2026don,tang2025dnsgreen,wu2025promptguidedduallatentsteering,huang2026machine} and has been successfully applied across various learning paradigm, including unsupervised learning~\cite{proto6}, supervised learning~\cite{proto2, proto4}, zero-shot~\cite{proto7}, and few-shot~\cite{proto8, proto9, proto10}.  
For instance, a prototype-based semantic segmentation approach~\cite{proto1} has been proposed to distinguish different semantic categories using representative prototypes to enhance model performance.  
In addition, another technique utilizes vector quantization~\cite{proto3} to construct a discrete latent representation, which benefits image reconstruction tasks.
Furthermore, Guo \textit{et al.}~\cite{proto10} proposed a lightweight embedding layer based on prototype loss, together with a joint fine-tuning strategy, to achieve efficient interactive segmentation of root-system images under conditions of scarce annotations. It has been shown to preserve intra-class variability and accelerate the user-annotation process.
These achievements give some insights into the use of prototypes to be able to learn accurate neighbor relations. 
\textcolor{revisecolor2}{However, traditional prototype methods often require explicit, pre-defined attribute labels (e.g., ``color'', ``shape''), which are unavailable in open-domain CIR datasets (e.g., CIRR). They lack the flexibility to adaptively disentangle latent attributes from arbitrary natural language queries, leaving a gap in applying prototype learning to open-ended retrieval tasks.}
Thus, we perform adaptive mining of prototypes and then utilize prototypes to calibrate the attribute distribution of a multimodal query.

%% file: 3_met.tex
\section{COMBINER}

\begin{figure*}[h]
\centering
	\includegraphics[width=\linewidth]{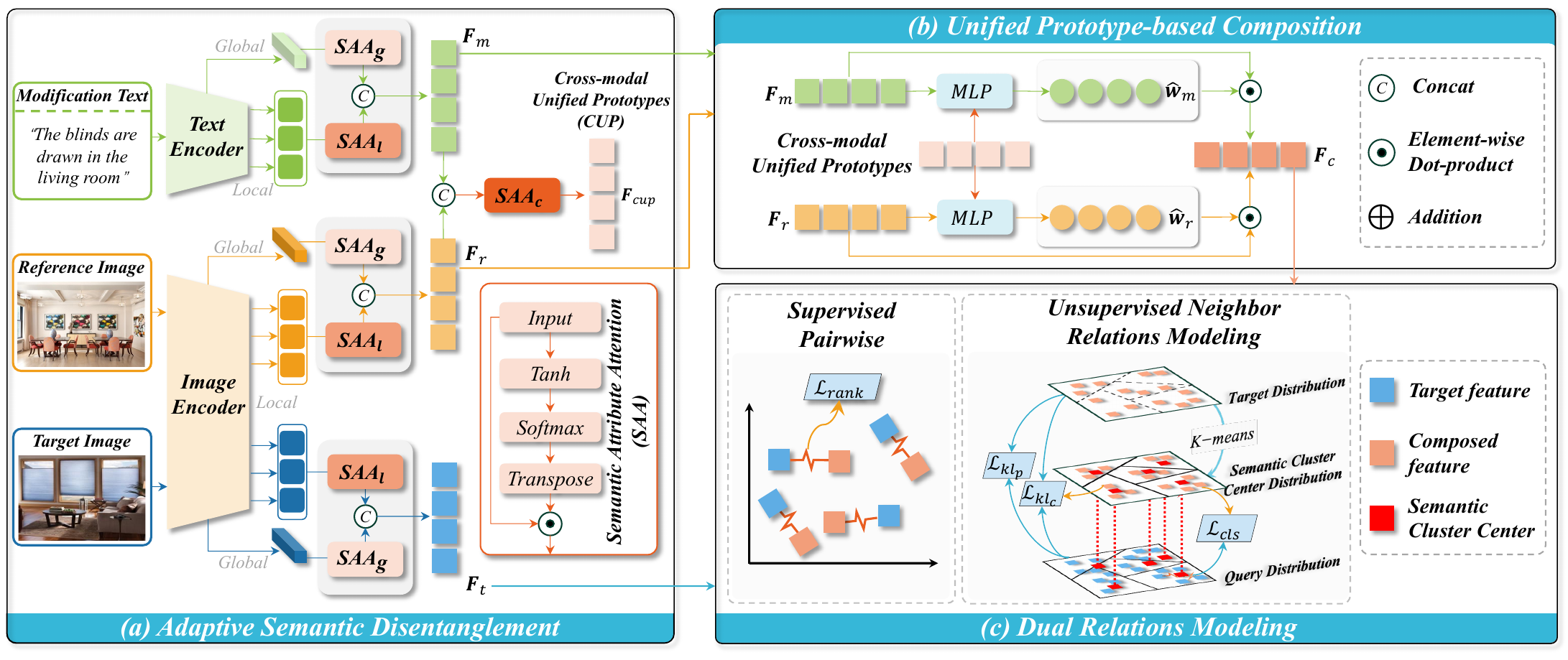}
        \vspace{-16pt}
	\caption{Overall framework of COMBINER, which consists of (a) Adaptive Semantic Disentanglement, (b) Unified Prototype-based Composition, and (c) Dual Relations Modeling.}
    \vspace{-20pt}
	\label{fig:Framework}
\end{figure*}
As a major novelty, our model aims to optimize the multimodal composition and metric learning via learning attribute-based neighbor relations. 
As illustrated in Fig.~\ref{fig:Framework}, COMBINER consists of three key modules: 
(a) \textit{Adaptive Semantic Disentanglement}, which utilizes \textit{Semantic Attribute Attention (SAA)} to adaptively disentangle the multi-grained semantic attribute prototypes of triplets;
(b) \textit{Unified Prototypes-based Composition}, which eliminates the effects of modal heterogeneity on attribute prototypes with the benefit of CUP for deriving attribute-level composed features;
And (c) \textit{Dual Relations Modeling}, which is designed to model supervised pairwise relations and unsupervised neighbor relations based on the attribute-level composed features. 
\textcolor{revisecolor}{We provide a \textit{Symbol Table} for the COMBINER architecture, as shown in Table~\ref{tab:Symbol_Table}.}

\subsection{Preliminary}
\label{sec:pre}
Our proposed COMBINER aims to tackle the composed image retrieval (CIR) task, which seeks target images matching a multimodal query composed of a reference image and a modification text. We assume a collection of $N$ training triplets $\mathcal{T}=\left\{\left(x_{r},t_{m}, x_{t}\right)_{n}\right\}_{n=1}^{N}$, where $x_{r}$ is the reference image, $t_{m}$ is the modification text, and $x_{t}$ refer to the desired target image. Our goal is to learn an embedding space in which the encoded multimodal query $\left(x_{r},t_{m}\right)$ is mapped as closely as possible to the embedding of its corresponding target $x_{t}$. Formally, we seek
\begin{equation}
    \mathcal{M}\left(x_r, t_m\right) \rightarrow \mathcal{M}\left(x_t\right),\label{eq1}
\end{equation}
where $\mathcal{M}$ denotes the learned embedding operation applied to both the multimodal query and the target image.

\subsection{Adaptive Semantic Disentanglement}
\label{sec:asd}

This component is designed to adaptively disentangle the attribute-prototype features associated with each element in a triplet, thereby establishing a solid basis for modeling neighbor relations. In particular, we adopt the contrastive language-vision pretrained network, CLIP~\cite{clip}, as our feature backbone to obtain both global and local features for each triplet. Prior studies have shown that leveraging both global and local CLIP features substantially benefits CIR performance~\cite{clip4cir-v2, clvcnet, tgcir}. Let $\hat{\textbf{F}}^g\in \mathbb{R}^{1 \times D}$, $\dot{\textbf{F}}^l\in \mathbb{R}^{S \times \dot{D}}$ respectively denote the features from the last and penultimate CLIP layer, where $S$ is the number of channels per sample, $D$ and $\dot{D}$ denote the global and local feature dimension of CLIP, respectively. For facilitating the concatenation between the global and local attribute prototype features, a linear transformation layer $\operatorname{FC}$ is initially employed, as formulated below,
\begin{equation}
\hat{\textbf{F}}^l=\operatorname{FC}(\dot{\textbf{F}}^l). \\
    \label{eq4}
\end{equation}

Subsequently, we design the \textit{Semantic Attribute Attention (SAA)}, which aims to capture the prototypes from original features to adaptively disentangle \mbox{attribute prototype} features. Specifically, we utilize \textit{SAA} for both granularities, denoted as $\operatorname{\textit{SAA}}_g$ and $\operatorname{\textit{SAA}}_l$, respectively, since both the global and local CLIP features entangle the attribute prototypes. \textit{SAA} module is detailed as follows.

\textit{\textbf{Semantic Attribute Attention (SAA).}}
\label{Semantic Category Attention}
\textcolor{revisecolor}{We recognize that the original features from CLIP are often entangled, i.e., a single vector blends multiple semantics such as color, shape, and texture. The purpose of \textit{SAA} is to address this issue. Specifically, \textit{SAA} acts as a set of ``learnable semantic filters'', as shown in Eq.(\ref{eq:eq3}). It learns a set of attention weights $\textbf{W}_c$ to disentangle and extract $Y$ (or $X$) independent attribute prototypes from the mixed original features (e.g., one prototype representing ``dog'', another representing ``bedding''). This step is crucial because it enables our model to compare samples at a fine-grained attribute level, rather than relying on a single, coarse-grained overall feature, thus more accurately identifying visually similar but attribute-irrelevant distractors.}

Given that both global and local features undergo identical processing and each triplet element is treated uniformly, we illustrate the procedure using the local feature of the reference image $x_r$, which is denoted as $\hat{\textbf{F}}_r^l \in \mathbb{R}^{S \times D}$. First, we adaptively disentangle at the prototype-level by capturing contextual semantics as follows,

\begin{equation}
\left\{
\begin{aligned}
    &\textbf{A}=\operatorname{Tanh}(\hat{\textbf{F}}_r^l\textbf{W}_h)\textbf{W}_m , \\
    &\textbf{W}_{c} = \operatorname{Softmax}(\textbf{A}^\top), \\
\end{aligned}
    \label{eq:eq3}
\right.
\end{equation}
where $\textbf{W}_h\in \mathbb{R}^{D \times D'}$, $\textbf{W}_m\in \mathbb{R}^{D' \times Y}$ $(\text{with }D'=D/2)$ represent the weights of the linear layers, $D$ denotes the embedding dimension, $Y$ indicates the total number of attribute prototypes to be disentangled, which can be set separately for local and global features, and $\textbf{A}$ can be viewed as the MLP output without bias. Finally, the prototype weights $\textbf{W}_c\in \mathbb{R}^{Y \times S}$ are derived by \mbox{softmax} and we \mbox{weighted-aggregate} $\hat{\textbf{F}}_r^l$ with $\textbf{W}_c$ to obtain the attribute prototype features $\textbf{F}_r^l\in \mathbb{R}^{Y \times D}$ of the reference image, formulated as follows,

\begin{equation}
\textbf{F}_r^l = \textbf{W}_{c} \hat{\textbf{F}}_r^l.
    \label{eq:eq4}
\end{equation}

Similar to the disentanglement process described in \textit{SAA}, we disentangle the attribute prototype features $\textbf{F}^g_r\in \mathbb{R}^{X \times D}$ of the reference image from its global feature, where $X$ is set as the number of global attribute prototypes. Finally, we concatenate $\textbf{F}^g_r$ and $\textbf{F}^l_r$ to construct the multi-grained attribute prototype features of the reference image, denoted as $\textbf{F}_r=[\textbf{F}^g_r, \textbf{F}^l_r]\in \mathbb{R}^{U \times D}$ $(U=X+Y)$. Analogously, we adaptively disentangle the multi-grained attribute prototype features for the modification text and target image, denoted as $\textbf{F}_m, \textbf{F}_t\in \mathbb{R}^{U \times D}$, via the same $\operatorname{\textit{SAA}}_g$ and $\operatorname{\textit{SAA}}_l$ for deriving $\textbf{F}_r$. 

\subsection{Unified Prototype-based Composition}
\label{sec:upc}
\textcolor{revisecolor}{Through \textit{SAA}, we have disentangled the attribute prototypes. However, merely disentangling the attributes is insufficient, because the image and text prototypes suffer from the problem of modal inconsistency. For example, the ``dog'' prototype extracted from the image and the ``lying position'' prototype extracted from the text exist in unaligned feature spaces.}
To mitigate the modal heterogeneity between the prototypes and obtain the unified representation of cross-modal features based on attribute prototypes, we introduce a \textit{Unified Prototype-based Composition} module. This module composes the attribute prototypes of elements in the multimodal query by enforcing alignment through a unified prototype criterion.
It first utilizes \textit{SAA} (detailed in Section~\ref{Semantic Category Attention}) to disentangle the \textit{Cross-modal Unified Prototypes (CUP)} from the simply concatenated multimodal query feature. 

Specifically, we concatenate the attribute prototype features from the reference image and the modification text to obtain the concatenation features, defined as $\ddot{\textbf{F}}_c=[\textbf{F}_r, \textbf{F}_m]\in \mathbb{R}^{2U \times D}$. Then, we disentangle the concatenation semantics in $\ddot{\textbf{F}}_c$ via \textit{SAA} to derive CUP, denoted as $\textbf{F}_{cup}\in \mathbb{R}^{U \times D}$,
\begin{equation}
\textbf{F}_{cup}=\operatorname{\textit{SAA}}_c(\ddot{\textbf{F}}_c),
    \label{eq5}
\end{equation}
where the number of attribute prototypes in $\operatorname{\textit{SAA}}_c$ is set to $U$. 
\textcolor{revisecolor3}{It is worth noting that while CUP utilizes an attention-like weighting mechanism, it is fundamentally distinct from standard attention-based composite methods (e.g., cross-attention) widely used in previous CIR literature. First, standard attention typically operates on entangled raw spatial or text tokens, which often leads to coarse feature blending. In contrast, CUP operates strictly on the disentangled attribute prototypes extracted by the ASD module, enabling fine-grained, algebraic-like compositional operations. Second, standard cross-attention directly projects image features into the text space (or vice versa), which struggles with the inherent modality gap. CUP avoids this by dynamically generating a shared cross-modal semantic dictionary ($\textbf{F}_{cup}$). By independently aligning both image and text prototypes to this neutral $\textbf{F}_{cup}$ anchor, our method effectively resolves modal heterogeneity before feature composition.}
Subsequently, we regard CUP as the concatenation criteria and utilize the linear learning capability of MLP to obtain the feature weights over attribute prototype features from both the reference image and the modification text. 
\textcolor{revisecolor}{We use this ``shared dictionary'' $\textbf{F}_{cup}$ as a ``query'' to respectively interrogate the image prototypes $\textbf{F}_r$ and the text prototypes $\textbf{F}_m$ (as shown in Eq.(\ref{eq:eq6})), in order to obtain two sets of weights, $\hat{\textbf{w}}_r$ and $\hat{\textbf{w}}_m$, which are aligned in the CUP space. These weights determine the semantic proportion of the unified attributes from the image and the text when constructing the final composed features.}
These weights are formulated as follows,

\begin{equation}
\left\{
\begin{aligned}
    &\hat{\textbf{w}}_r=\sigma(\operatorname{MLP}([\textbf{F}_{cup}, \textbf{F}_r])),\\
    &\hat{\textbf{w}}_m=\sigma(\operatorname{MLP}([\textbf{F}_{cup}, \textbf{F}_m])), \\
\end{aligned}
    \label{eq:eq6}
\right.
\end{equation}
where $\sigma$ is the sigmoid function, $\hat{\textbf{w}}_r\in \mathbb{R}^{U}$ and $\hat{\textbf{w}}_m\in \mathbb{R}^{U}$ denote the feature weights relative to CUP, respectively. 
Finally, we aggregate the original attribute prototype features with above feature weights to obtain the attribute-level composed features of multimodal query $\textbf{F}_c\in \mathbb{R}^{U \times D}$ as follows,

\begin{equation}
\textbf{F}_c=\hat{\textbf{w}}_r \odot \textbf{F}_r + \hat{\textbf{w}}_m \odot \textbf{F}_m.
    \label{eq:eq7}
\end{equation}

For convenience, in the latter section, we refer to the attribute-level composed features of multimodal query simply as \textit{composed features}, denoted as $\textbf{F}_c$, and attribute prototype features of the target image as \textit{target features}, denoted as $\textbf{F}_t$.

\subsection{Dual Relations Modeling}
\label{sec:drm}
This module intends to model supervised pairwise relations and unsupervised neighbor relations by leveraging attribute-level features. The subsequent section outlines the details of the modeling process.

\subsubsection{Supervised Pairwise Relations Modeling}
To learn the pairwise relations, we adopt the batch-based classification loss~\cite{batch-based-classification-loss}, which is frequently employed in CIR tasks. This loss encourages each composed feature to move closer to its matching target features in the learned metric space.
Notably, since the acquisition of composed features relies on CUP, the process of modeling pairwise relations also enables CUP to implicitly approach to the target features.
Specifically, based on the average pooled composed feature and target feature, denoted as $\bar{\textbf{F}}_c$ and $\bar{\textbf{F}}_t$, respectively, we define the following batch-based classification loss,
\begin{equation}
    \mathcal{L}_{rank} = \frac{1}{B} \sum_{i=1}^{B} -\log \left\{ \frac{\exp \left\{ \operatorname{s} \left( \bar{\textbf{F}}_{ci} , \bar{\textbf{F}}_{ti} \right)  / \tau\right\}}{ \sum_{j=1}^{B} \exp \left\{ \operatorname{s} \left( \bar{\textbf{F}}_{ci}, \bar{\textbf{F}}_{tj} \right) / \tau \right\}  } \right\},
    \label{eq:eq8}
\end{equation}
where $\bar{\textbf{F}}_{ci}\in \mathbb{R}^{D}$ and $\bar{\textbf{F}}_{ti}\in \mathbb{R}^{D}$ represent average pooled composed features and target features of the $i$-th triplet, respectively. $B$ refers to the mini-batch size, $s(\cdot, \cdot)$ represents the cosine similarity computation, and $\tau$ stands for the temperature scaling factor. In the following formulas, the symbols consistent with this formula denote the same meanings.

\subsubsection{Unsupervised Neighbor Relations Modeling}
\label{sec:udrm}
To identify visually similar but attribute-unrelated images, we further propose an \textit{Unsupervised Neighbor Relations Modeling} based on attribute prototypes, which evaluates the attribute relevance among samples using attribute similarity assessments. Based on this, it employs unsupervised clustering to gather attribute-related images while pushing away attribute-unrelated images to optimize metric learning.

Specifically, we first utilize the \textit{K-means} clustering algorithm on the average pooled target attribute prototype features $\bar{\textbf{F}}_t$ to unsupervisedly divide them into $H$ semantic clusters, and obtain $H$ semantic cluster centroids, which jointly form the semantic centroid set $\mathcal{P}=\{p_1,...,p_h,..., p_H\}$, where $H$ denotes the number of semantic cluster centroids. For each triplet, the corresponding semantic cluster centroid can be treated as a common visually similar and attribute-related neighbor of both the multimodal query and the target image, since they are relatively close in the cluster. Through the semantic clustering process, the visually similar and attribute-related images can be gathered in the metric space for better learning of the neighbor relations while the attribute-unrelated images are pushed away. Thus, we reorganize each training triplet data into neighbor relations triplets (\textit{i.e.}, the composed feature, the target feature, and their semantic cluster center features) and all the neighbor relations triplets collectively form a set $\mathcal{D}\!\!=\!\!\left\{\left(\bar{\textbf{F}}_c,\bar{\textbf{F}}_t,\bar{\textbf{F}}_p\right)_{n}\right\}_{n=1}^{N}$, where $N$ is training triplet data number, $\bar{\textbf{F}}_p$ is the corresponding semantic cluster centroid feature, $p\!\in\!\mathcal{P}$ and $|\mathcal{P}|\!=\!H$. For convenience, all subsequent ``triplet'' in this section denote the neighbor relation triplet.

\begin{table*}[h]
  \centering
  \tabcolsep=12pt
    \renewcommand\arraystretch{1.2}
      \caption{Performance comparison on the FashionIQ dataset with respect to R@$k$ ($\%$). The top-performing results across all methods are highlighted in \textcolor{defblue}{blue}, and the best results among baseline methods are indicated with underlining.}
      \vspace{-5pt}
      \resizebox{0.85\linewidth}{!}{
    \begin{tabular}{l|cc|cc|cc|cc}
    \hline
    \hline    \multicolumn{1}{c|}{\multirow{2}{*}{Method}}     & \multicolumn{2}{c|}{Dresses} & \multicolumn{2}{c|}{Shirts} & \multicolumn{2}{c|}{Tops\&Tees} & \multicolumn{2}{c}{Avg}  \\
\cline{2-9}         & R@$10$  & R@$50$  & R@$10$  & R@$50$  & \multicolumn{1}{c}{R@$10$} & R@$50$  & R@$10$  & R@$50$   \\ \hline 
\rowcolor[rgb]{ .949,  .949,  .949} \multicolumn{9}{c}{\textit{Traditional Model-Based Methods}}\\
    TIRG~\cite{tirg}~\textcolor{gray}{(CVPR'19)} & 14.87  & 34.66  & 18.26  & 37.89  & 19.08  & 39.62  & 17.40  & 37.39   \\
    VAL~\cite{val}~\textcolor{gray}{(CVPR'20)}& 21.12  & 42.19  & 21.03  & 43.44  & 25.64  & 49.49  & 22.60  & 45.04    \\
    CIRPLANT~\cite{cirplant}~\textcolor{gray}{(ICCV'21)}& 17.45  & 40.41  & 17.53  & 38.81  & 21.64  & 45.38  & 18.87  & 41.53 \\
    ARTEMIS~\cite{artemis}~\textcolor{gray}{(ICLR'22)} & 27.16  & 52.40  & 21.78  & 43.64  & 29.20  & 54.83  & 26.05  & 50.29   \\
    EER~\cite{eer}~\textcolor{gray}{(TIP'22)}& 30.02  & 55.44  & 25.32  & 49.87  & 33.20  & 60.34  & 29.51  & 55.22   \\
    CRN~\cite{crn}~\textcolor{gray}{(TIP'23)}& 32.67   & 59.30   & 30.27   & 56.97   & 37.74  & 65.94  & 33.56  & 60.74     \\
    MGUR~\cite{mgur}~\textcolor{gray}{(ICLR'24)}& 32.61  & 61.34  & 33.23  & 62.55  & 41.40  & 72.51  & 35.75  & 65.47    \\
\hdashline
\rowcolor[rgb]{ .949,  .949,  .949} \multicolumn{9}{c}{\textit{VLP Model-Based Methods}}\\
    LF-CLIP~\cite{lfclip}~\textcolor{gray}{(CVPR'22)}& 31.63  & 56.67  & 36.36  & 58.00  & 38.19  & 62.42  & 35.39  & 59.03  \\
    Prog. Lrn.~\cite{Prog-Lrn}~\textcolor{gray}{(SIGIR'22)}& 38.18  & 64.50  & 48.63  & 71.54  & 52.32  & 76.90 & 46.38  & 70.98   \\
    FAME-ViL~\cite{famevil}~\textcolor{gray}{(CVPR'23)} & 42.19  & 67.38  & 47.64  & 68.79  & 50.69  & 73.07  & 46.84  & 69.75  \\
    TG-CIR~\cite{tgcir}~\textcolor{gray}{(ACM MM'23)}  & 45.22  & 69.66  & 52.60  & 72.52  & 56.14 & 77.10  & 51.32  & 73.09 \\
    SyncMask~\cite{syncmask}~\textcolor{gray}{(CVPR'24)}& 33.76  & 61.23  & 35.82  & 62.12  & 44.82  & 72.06  & 38.13  & 65.14  \\
    AFCE~\cite{afce}~\textcolor{gray}{(TIP'24)}& 33.98 & 59.96 & 40.15 & 62.76 & 43.75 & 60.70 & 39.29 & 63.47\\
    SADN~\cite{SADN}~\textcolor{gray}{(ACM MM'24)} &40.01 & 65.10 & 43.67 & 66.05 & 48.04 & 70.93 & 43.91 & 67.36 \\
    CoVR-2~\cite{covr-2}~\textcolor{gray}{(TPAMI'24)} &46.53& 69.60 &51.23 &70.64& 52.14 &73.27& 49.96 &71.17\\
    Candidate~\cite{candidate}~\textcolor{gray}{(TMLR'24)}& 48.14& 71.34& 50.15 &71.25& 55.23& 76.80& 51.17& 73.13\\
    SPRC~\cite{sprc}~\textcolor{gray}{(ICLR'24)}& 49.18 &	72.43 &	55.64 &	73.89 	&59.35 &	78.58 &	54.72 &	74.97 \\
    LIMN~\cite{limn}~\textcolor{gray}{(TPAMI'24)} & 50.72  & 74.52  & 56.08  & 77.09  & 60.94  & 81.85  & 55.91 & 77.82 \\
    LIMN+~\cite{limn}~\textcolor{gray}{(TPAMI'24)} &52.11 &75.21& 57.51& 77.92& 62.67& 82.66&57.43 &	78.60  \\
    DQU-CIR~\cite{dqu}~\textcolor{gray}{(SIGIR'24)}& \underline{57.63} &	\underline{78.56} 	&\underline{62.14} &	\underline{80.38} &	\underline{66.15} &	\underline{85.73} &	\underline{61.97} &	\underline{81.56} \\
    IUDC~\cite{iudc}~\textcolor{gray}{(TOIS'25)}& 35.22& 61.90& 41.86& 63.52& 42.19& 69.23& 39.76& 64.88  \\
    DIPNEC~\cite{dipnec}~\textcolor{gray}{(AAAI'25)}& 46.90& 71.29& 56.92& 77.77& 58.18& 80.88& 54.00& 76.64  \\
    ENCODER~\cite{encoder}~\textcolor{gray}{(AAAI'25)}& 51.51& 76.95& 54.86& 74.93& 62.01& 80.88& 56.13& 77.59  \\

\rowcolor{gray!15}
    \multicolumn{1}{l|}{\textbf{COMBINER(Ours)}} & \textcolor{defblue}{\textbf{57.96}} &	\textcolor{defblue}{\textbf{79.08}} &	\textcolor{defblue}{\textbf{63.64}} &	\textcolor{defblue}{\textbf{81.84}} &	\textcolor{defblue}{\textbf{68.18}} &	\textcolor{defblue}{\textbf{85.87}} &	\textcolor{defblue}{\textbf{63.26}} &	\textcolor{defblue}{\textbf{82.27}}     \\
    \hline    \hline
    \end{tabular}
}
\vspace{-19pt}
  \label{tab:fiq_shoes}%
\end{table*}%

\setlength{\textfloatsep}{0pt}
\begin{algorithm}[h]
\small
\caption{\textcolor{revisecolor}{Training Procedure of COMBINER}}
\label{alg:combiner}
\SetKwInOut{Input}{Input}
\SetKwInOut{Output}{Output}
\SetKwFunction{SAA}{SAA}
\SetKwFunction{Clip}{CLIP}
\SetKwFunction{KMeans}{K-Means}

\Input{
    Training triplets $\mathcal{T} = \{(x_r, t_m, x_t)\}_{n=1}^N$; \\
    Pre-trained CLIP model; \\
    Hyper-parameters: $\tau, \rho, \kappa, \mu$; \\
    Attribute prototype number $U$, Cluster number $H$.
}
\Output{Trained COMBINER parameters $\Theta$.}

Initialize model parameters $\Theta$\;
\While{not converged}{
    Sample a mini-batch of triplets $\mathcal{B} = \{(x_r, t_m, x_t)\}_{i=1}^B$ from $\mathcal{T}$\;
    
    \textit{1. Feature Extraction}\;
    Extract global and local features using \Clip for $(x_r, t_m, x_t)$\;
    
    \textit{2. Adaptive Semantic Disentanglement (ASD)}\;
    Obtain attribute prototype features $\textbf{F}_r, \textbf{F}_m, \textbf{F}_t$ via \SAA (Eq.(\ref{eq:eq3})~(\ref{eq:eq4}))\;
    
    \textit{3. Unified Prototype-based Composition (UPC)}\;
    Concatenate features: $\dot{\textbf{F}}_c \leftarrow [\textbf{F}_r, \textbf{F}_m]$\;
    Generate Cross-modal Unified Prototypes (CUP): $\textbf{F}_{cup} \leftarrow \SAA(\dot{\textbf{F}}_c)$ (Eq.(\ref{eq5}))\;
    Compute composed feature weights $\hat{\textbf{w}}_r, \hat{\textbf{w}}_m$ relative to $\textbf{F}_{cup}$ (Eq.(\ref{eq:eq6}))\;
    Construct composed features: $\textbf{F}_c \leftarrow \hat{\textbf{w}}_r \odot \textbf{F}_r + \hat{\textbf{w}}_m \odot \textbf{F}_m$ (Eq.(\ref{eq:eq7}))\;
    
    \textit{4. Dual Relations Modeling (DRM)}\;
    \textbf{Pairwise:} Compute supervised batch-based classification loss $\mathcal{L}_{rank}$ (Eq.(\ref{eq:eq8}))\;
    
    \textbf{Neighbor:} \\
    Update semantic centroids $\mathcal{P} = \{p_1, \dots, p_H\}$ via \KMeans on target features $\textbf{F}_t$\;
    Compute cluster-oriented classification loss $\mathcal{L}_{cls}$ (Eq.(\ref{eq9}))\;
    Compute distribution consistency losses $\mathcal{L}_{kl_c}$ and $\mathcal{L}_{kl_p}$ (Eq.(\ref{eq11})~(\ref{eq12}))\;
    
    \textit{5. Optimization}\;
    Calculate total loss: $\mathcal{L}_{total} = \mathcal{L}_{rank} + \rho\mathcal{L}_{cls} + \kappa\mathcal{L}_{kl_p} + \mu\mathcal{L}_{kl_c}$ (Eq.(\ref{eq13}))\;
    Update $\Theta$ via back-propagation to minimize $\mathcal{L}_{total}$\;
}
\Return{$\Theta$}
\end{algorithm}

To explicitly gather the attribute-related elements in the triplets, we promote both the multimodal query and the corresponding target image to approach their associated semantic cluster centroids in the corresponding semantic clusters by the \textit{cluster-oriented batch-based classification} loss function $\fontsize{9pt}{9pt}\mathcal{L}_{cls}\!\!=\!\mathcal{L}_{cls}^c\!+\!\mathcal{L}_{cls}^t$, where $\fontsize{9pt}{9pt}\mathcal{L}_{cls}^c$ and $\fontsize{9pt}{9pt}\mathcal{L}_{cls}^t$ are formulated as,
\begin{equation}
\left\{
\begin{aligned}
    &\mathcal{L}_{cls}^c = \frac{1}{B} \sum_{i=1}^{B} -\log \left\{ \frac{\exp \left\{ \operatorname{s} \left( \bar{\textbf{F}}_{ci} , \bar{\textbf{F}}_{pi} \right)  / \tau\right\}}{ \sum_{j=1}^{B} \exp \left\{ \operatorname{s} \left(\bar{\textbf{F}}_{ci}, \bar{\textbf{F}}_{pj} \right) / \tau \right\}  } \right\}, \\
    &\mathcal{L}_{cls}^t = \frac{1}{B} \sum_{i=1}^{B} -\log \left\{ \frac{\exp \left\{ \operatorname{s} \left( \bar{\textbf{F}}_{ti} , \bar{\textbf{F}}_{pi} \right)  / \tau\right\}}{ \sum_{j=1}^{B} \exp \left\{ \operatorname{s} \left( \bar{\textbf{F}}_{ti}, \bar{\textbf{F}}_{pj} \right) / \tau \right\}  } \right\}. \\
\end{aligned}
\right.
    \label{eq9}
\end{equation}

Moreover, to make the attribute-related multimodal query and the target image as close as possible to each other in terms of attribute distribution, we argue that the similarity distributions of their attribute-level features with the corresponding cluster centroid features should also be consistent in the metric space. 
Inspired by~\cite{clvcnet}, we utilize the mutual learning strategy to promote the similarity distributions to be consistent. We first compute the similarity distribution between each multimodal query and the semantic cluster centroids. Specifically, assuming that there are $B$ triplets in each batch, we define $\mathbf{p}_i^{c}=[p_{i1}^{c},...,p_{iB}^{c}]$ as the batch-wise similarity distribution corresponding to the $i$-th composed feature, where $p_{ij}^{c}$ quantifies the similarity between the $i$-th composed feature and the cluster center of the $j$-th triplet, formulated as follows,

\begin{equation}
p_{ij}^{c} =  \frac{\exp \left\{ \operatorname{s} \left( \bar{\textbf{F}}_{ci} , \bar{\textbf{F}}_{pj} \right)  / \tau\right\}}{ \sum_{b=1}^{B} \exp \left\{ \operatorname{s} \left(\bar{\textbf{F}}_{ci}, \bar{\textbf{F}}_{pb} \right) / \tau \right\}  }. 
\label{eq10}
\end{equation}

Analogously, we can derive the similarity distribution $\mathbf{p}_i^{t}=[p_{i1}^{t},...,p_{iB}^{t}]$ of the target feature with corresponding cluster center features. Subsequently, we define \textit{\mbox{cluster-oriented} similarity distribution regularization} via Kullback Leibler (KL) Divergence to promote the consistency between $\mathbf{p}_i^{t}$ and $\mathbf{p}_i^{c}$, formulated as follows,
\begin{equation}
\mathcal{L}_{kl_c}= \frac{1}{B}\sum_{i=1}^{B}D_{KL}\left({\mathbf{p}}_i^{c} \| {\mathbf{p}}_i^{t}\right)= \frac{1}{B}\sum_{i=1}^{B}\sum_{j=1}^{B} p_{ij}^{c} \log \frac{p_{ij}^{c}}{p_{ij}^{t}}.
\label{eq11}
\end{equation}

Additionally, to mitigate the attribute semantic entanglement caused by the average pool of composed features and target features, we design \textit{pool-oriented distribution consistency regularization} to pull their similarity distributions to be consistent. Similarly to Eqn.~(\ref{eq12})~(\ref{eq13}), we can obtain the similarity distribution $\mathbf{p}_i^{ct}=[p_{i1}^{ct},...,p_{iB}^{ct}]$, where $p_{ij}^{ct}$ refers to the similarity between the $i$-th average pooled composed features and the $j$-th average pooled target features, and $\mathbf{p}_i^{tt}=[p_{i1}^{tt},...,p_{iB}^{tt}]$, where $p_{ij}^{tt}$ represents the similarity between the $i$-th target features and the $j$-th target features. Then we formulate the \textit{pool-oriented distribution consistency regularization} with KL Divergence as follows,
\begin{equation}
\mathcal{L}_{kl_p}= \frac{1}{B}\sum_{i=1}^{B}D_{KL}\left({\mathbf{p}}_i^{tt} \| {\mathbf{p}}_i^{ct}\right)= \frac{1}{B}\sum_{i=1}^{B}\sum_{j=1}^{B} p_{ij}^{tt} \log \frac{p_{ij}^{tt}}{p_{ij}^{ct}}.
\label{eq12}
\end{equation}

Finally, we obtain the metric space optimization function for COMBINER as follows,
\begin{equation}
\mathbf{\Theta^{*}}=
\underset{\mathbf{\Theta}}{\arg \min }( {\mathcal{L}}_{rank} + \rho \mathcal{L}_{cls} + \kappa \mathcal{L}_{kl_p} + \mu \mathcal{L}_{kl_c}),
\label{eq13}
\end{equation}
where $\mathbf{\Theta}$ denotes the set of learnable parameters in COMBINER, while $\rho, \kappa, \mu$ serve as trade-off hyper-parameters.
\textcolor{revisecolor}{To provide a clearer understanding of COMBINER's architectural flow, we illustrate its training procedure in Algorithm~\ref{alg:combiner}.}

%% file: 4_exp.tex
\section{Experiment}

\begin{table*}[ht]
  \centering
  \vspace{-5pt}
  \tabcolsep=12pt
    \renewcommand\arraystretch{1.2}
      \caption{Performance comparison on CIRR with respect to R@$k$($\%$) and R$_{subset}$@$k$($\%$). The overall best results are colored in \textcolor{defblue}{blue}, while the best results over baselines are underlined.}
      \vspace{-5pt}
      \resizebox{0.86\linewidth}{!}{
    \begin{tabular}{l|cccc|ccc|c}
\hline\hline
    \multicolumn{1}{c|}{\multirow{2}{*}{Method}} & \multicolumn{4}{c|}{R@$k$}      & \multicolumn{3}{c|}{R$_{subset}$@$k$} & \multirow{2}{*}{(R@5+R$_{subset}$@$1$)/$2$} \\
\cline{2-8}         & k=1   & k=5   & k=10  & k=50  & k=1   & k=2   & k=3   &  \\
    \hline
    \rowcolor[rgb]{ .949,  .949,  .949} \multicolumn{9}{c}{\textit{Traditional Model-Based Methods}} \\
    TIRG~\cite{tirg}~\textcolor{gray}{(CVPR'19)}& 14.61  & 48.37  & 64.08  & 90.03  & 22.67  & 44.97  & 65.14  & 35.52  \\
    CIRPLANT~\cite{cirplant}~\textcolor{gray}{(ICCV'21)}& 19.55  & 52.55  & 68.39  & 92.38  & 39.20  & 63.03  & 79.49  & 45.88  \\
    ARTEMIS~\cite{artemis}~\textcolor{gray}{(ICLR'22)}& 16.96  & 46.10  & 61.31  & 87.73  & 39.99  & 62.20  & 75.67  & 43.05  \\
    ComqueryFormer~\cite{comquery-former}~\textcolor{gray}{(TMM'23)}& 25.76  & 61.76  & 75.90  & 95.13  & 51.86  & 76.26  & 89.25  & 56.81  \\
    \hdashline
\rowcolor[rgb]{ .949,  .949,  .949} \multicolumn{9}{c}{\textit{VLP Model-Based Methods}}\\
    LF-CLIP~\cite{lfclip}~\textcolor{gray}{(CVPR'22)}& 33.59  & 65.35  & 77.35  & 95.21  & 62.39  & 81.81  & 92.02  & 63.87  \\
    TG-CIR~\cite{tgcir}~\textcolor{gray}{(ACM MM'23)} & 45.25  & 78.29  & 87.16  & 97.30  & 72.84  & 89.25  & 95.13  & 75.57  \\
    AFCE~\cite{afce}~\textcolor{gray}{(TIP'24)}& 39.80& 74.24 & 85.71 & 97.23 & 69.43 & 86.93 & 94.14 & 71.84\\
    LIMN~\cite{limn}~\textcolor{gray}{(TPAMI'24)} &43.64  & 75.37  & 85.42  & 97.04  & 69.01  & 86.22  & 94.19  & 72.19 \\
    LIMN+~\cite{limn}~\textcolor{gray}{(TPAMI'24)} &43.33 &	75.41 &	85.81& 	97.21 &	69.28 &	86.43 &	94.26 	&72.35 \\
    SSN~\cite{ssn}~\textcolor{gray}{(AAAI'24)}& 43.91  & 77.25  & 86.48  & 97.45 & 71.76  & 88.63  & 95.54  & 74.51  \\
    SADN~\cite{SADN}~\textcolor{gray}{(ACM MM'24)} &44.27& 78.10& 87.71 &97.89& 72.71 &89.33& 95.38& 75.41\\
    DQU-CIR~\cite{dqu}~\textcolor{gray}{(SIGIR'24)}& 46.22 &	78.17 &	87.64 &	97.81 	&70.92 &	87.69 &	94.68 &	74.55  \\
    CoVR-2~\cite{covr-2}~\textcolor{gray}{(TPAMI'24)} &50.43 &81.08& 88.89& \underline{98.05}& 76.75& 90.34& 95.78 & 79.28\\
    Candidate~\cite{candidate}~\textcolor{gray}{(TMLR'24)}& 50.55 &	81.75 &	\underline{89.78} &	97.18 &	80.04 &	91.90 &	96.58 &	80.90 \\
    SPRC~\cite{sprc}~\textcolor{gray}{(ICLR'24)}& 51.96 &	82.12 &	89.74 &	97.69 &	\underline{80.65} &	\underline{92.31} &	\underline{96.60} 	& \underline{81.39}  \\
    ENCODER~\cite{encoder}~\textcolor{gray}{(AAAI'25)}& 46.10& 77.98& 87.16 &97.64& 76.92& 90.41& 95.95 &77.45\\
    DIPNEC~\cite{dipnec}~\textcolor{gray}{(AAAI'25)}& 47.24& 80.20& 89.07 &97.87& 73.97& 89.74& 95.72 &77.09\\
   \rowcolor{gray!15}
    \multicolumn{1}{l|}{\textbf{COMBINER(Ours)}}  &     \textcolor{defblue}{\textbf{52.60}} &	\textcolor{defblue}{\textbf{82.51}} &	\textcolor{defblue}{\textbf{90.12}} &	\textcolor{defblue}{\textbf{98.17}} &	\textcolor{defblue}{\textbf{81.33}} &	\textcolor{defblue}{\textbf{93.06}} &	\textcolor{defblue}{\textbf{97.23}}  & \textcolor{defblue}{\textbf{81.92}} \\
\hline \hline
    \end{tabular}
}
\vspace{-16pt}
  \label{tab:cirr}%
\end{table*}%

\begin{table}[ht]
\centering
    \renewcommand\arraystretch{1.2}
    \caption{Performance comparison on Shoes with respect to R@$k$($\%$). The overall best results are colored in \textcolor{defblue}{blue}, while the best results over baselines are underlined.}
    \vspace{-8pt}
    \tabcolsep=12pt
    \resizebox{\linewidth}{!}{
       \begin{tabular}{l|ccc|c}
\hline\hline
    \multicolumn{1}{c|}{Method} & R@$1$   & R@$10$  & R@$50$  & Avg \\
    \hline
    \rowcolor[rgb]{ .949,  .949,  .949} \multicolumn{5}{c}{\textit{Traditional Model-Based Methods}}\\
    TIRG~\cite{tirg}~\textcolor{gray}{(CVPR'19)}& 12.60  & 45.45  & 69.39  & 42.48  \\
    VAL~\cite{val}~\textcolor{gray}{(CVPR'20)}  & 17.18  & 51.52  & 75.83  & 48.18  \\
    EER~\cite{eer}~\textcolor{gray}{(TIP'22)}& 20.05 & 56.02  & 79.94  & 52.00 \\
    CRN~\cite{crn}~\textcolor{gray}{(TIP'23)}& 18.92  & 54.55  & 80.04  & 51.17 \\
    MGUR~\cite{mgur}~\textcolor{gray}{(ICLR'24)}& 18.41  & 53.63  & 79.84  & 50.63  \\
    \hdashline
    \rowcolor[rgb]{ .949,  .949,  .949} \multicolumn{5}{c}{\textit{VLP Model-Based Methods}}\\
    Prog. Lrn.~\cite{Prog-Lrn}~\textcolor{gray}{(SIGIR'22)}&  22.88  & 58.83  & 84.16  & 55.29  \\
    TG-CIR~\cite{tgcir}~\textcolor{gray}{(ACM MM'23)} & 25.89 &63.20 &85.07 &58.05 \\
    AFCE~\cite{afce}~\textcolor{gray}{(TIP'24)}& 19.10 & 44.31 & 55.37 & 79.57 \\
    LIMN~\cite{limn}~\textcolor{gray}{(TPAMI'24)} &- & 68.20 & 87.45 & -\\
    LIMN+~\cite{limn}~\textcolor{gray}{(TPAMI'24)} &- & 68.37 & 88.07 & -\\
    DQU-CIR~\cite{dqu}~\textcolor{gray}{(SIGIR'24)}& \underline{31.47} & \underline{69.19} & \underline{88.52} & \underline{63.06} \\
    IUDC~\cite{iudc}~\textcolor{gray}{(TOIS'25)}& 21.17 &56.82& 82.25& 53.41\\
    ENCODER~\cite{encoder}~\textcolor{gray}{(AAAI'25)}& 26.97 &65.59& 86.48& 59.68\\
    \rowcolor{gray!15}
    \multicolumn{1}{l|}{\textbf{COMBINER(Ours)}} & \textcolor{defblue}{\textbf{32.37}} &	\textcolor{defblue}{\textbf{70.81}} &	\textcolor{defblue}{\textbf{88.81}}& \textcolor{defblue}{\textbf{64.00}}  \\  	 	
\hline\hline
    \end{tabular}
}
      \label{tab:shoes}%
\end{table}%

In this section, we outline our experimental configuration and then conduct an in-depth analysis of the experiment by answering the following questions:

\begin{itemize}[leftmargin=8pt]
	\item \textbf{RQ1}: Does our proposed model COMBINER surpass the existing \mbox{state-of-the-art} methods?
	\item \textbf{RQ2}: How does each component affect COMBINER?
	\item \textbf{RQ3}: How does the hyper-parameters impact COMBINER?
	\item \textbf{RQ4}: How does COMBINER perform in terms of inference efficiency?
	\item \textbf{RQ5}: How is the qualitative performance of COMBINER?
\end{itemize}

\subsection{Experimental Settings}
\label{exp:experimental_settings}
\subsubsection{Datasets}
Following previous works, we selected two fashion domain datasets, Shoes~\cite{shoes}, FashionIQ~\cite{FashionIQ}, and an open domain dataset CIRR~\cite{cirr} for a fair comparison.

\subsubsection{Implementation Details}
Following previous work~\cite{dqu}, we utilize employ the ViT-H/14 variant of CLIP as the feature extraction backbone for \mbox{COMBINER}, and its embedding dimension $D$ is $1,024$. The global and local attribute prototype numbers $X,Y$ are both set to $4$, thereby $U=8$. The number of semantic clusters $H$ is fixed to $1,900$ for all three datasets. The three non-negative trade-off hyper-parameters were all searched using a grid search, and finally confirmed as $\rho=1.6, \kappa=0.5, \mu=0.5$. COMBINER is optimized with the AdamW optimizer, using an initial learning rate of $1e$-$4$ and a CLIP learning rate of $1e$-$6$. All experiments run for $10$ epochs with a batch size of $32$ on an NVIDIA A$40$ GPU equipped with $48$ GB of memory.

\subsubsection{Evaluation} 
For a fair comparison, we follow each dataset's standard evaluation protocol and report Recall@$k$ (R@$k$ for short). For Shoes, we reported R@$k$ ($k=1,10,50$) and their average values. For FashionIQ, we reported R@$10$, R@$50$, and their average values for each category. For CIRR, we reported R@$k$ ($k=1,5,10,50$), R$_{subset}$@$k$ ($k=1,2,3$), and the mean of R@$5$ and R$_{subset}$@$1$.

\subsection{Performance Comparison (RQ1)}
To validate COMBINER's performance, we compare it against established methods, such as TIRG~\cite{tirg}, CIRPLANT~\cite{cirplant} with traditional models (ResNet, LSTM, \textit{etc.}) as the backbone, and baselines such as DQU-CIR~\cite{dqu}, SPRC~\cite{sprc} based on the VLP model for comparison. Results on the FashionIQ, CIRR, and Shoes datasets are detailed in TABLES~\ref{tab:fiq_shoes}, \ref{tab:cirr}, and \ref{tab:shoes}, respectively.
From these in-depth results, we obtain the following insights.
\textbf{1)} COMBINER consistently exceeds every baseline across all three benchmarks, covering both fashion‐domain and open‐domain CIR datasets. Specifically, COMBINER achieves $2.1$\% for FashionIQ-Avg-R@$10$, $2.9$\% relative improvements for Shoes-R@$1$, and $1.2$\% for CIRR-R@$1$, which demonstrates the validity of COMBINER that learns attribute-based neighbor relations to optimize the metric space in both open domain and fashion-specific CIR datasets.
\textbf{2)} SPRC achieves sub-optimal performance on a majority of metrics on the open-domain dataset CIRR (such as R@$1$ and R$_{subset}$@$1$), but does not have a significant performance gain in the fashion-domain dataset FashionIQ (such as R@$10$ on Dresses). On the contrary, COMBINER consistently achieves optimal performance on all metrics. This indicates the superior retrieval capability of COMBINER which utilizes attribute distribution to model neighbor relations for the CIR task.
\textbf{3)} The improvement of COMBINER on Shoes and FashionIQ is more significant than that on CIRR, which may be due to the fact that the two fashion domain datasets have more visual similarity samples than the open domain dataset CIRR. 
For example, all of the images in the FashionIQ-Dresses dataset are dresses, and when two dresses share the same color, they can easily be treated as visually similar samples. And similar patterns can also be observed in the other fashion-domain datasets.
Thus, modeling the neighbor relations can lead to a greater performance improvement since there are more visually similar samples in fashion-domain datasets.

\begin{table}[h]
  \centering
  \renewcommand\arraystretch{1.2}
    \vspace{-7pt}
  \caption{\textcolor{revisecolor}{Performance comparison with ADDE-M on the CIRR dataset. \textbf{Bold} indicates the best result.}}
  \vspace{-4pt}
  \tabcolsep=1pt
  \resizebox{0.98\linewidth}{!}{
    \begin{tabular}{l|c|c|c}
      \hline
      \hline
      \multicolumn{1}{c|}{Method} & R@$1$ & R$_{subset}$@$1$ & Notes \\
      \hline
      ADDE-M~(Adjusted version) & 35.27\% & 68.14\% & Relies on pseudo attribute labels \\
      \rowcolor{gray!15}
      \textbf{COMBINER（Ours）} & \textbf{52.60\%} & \textbf{81.33\%} & \textbf{Dynamic prototypes (no explicit labels required)} \\
      \hline
      \hline
    \end{tabular}
  }
  \vspace{-12pt}
  \label{tab:ADDE-M}
\end{table}

{\textcolor{revisecolor}{\textbf{\textit{Comparison with the Attribute-based CIR model.}}}}
\textcolor{revisecolor}{To further demonstrate the superiority and effectiveness of the attribute semantic modeling method employed by COMBINER in processing general scenarios, we conduct a comparison with the representative Attribute-based CIR model ADDE-M~\cite{R-5} on the open-domain CIR dataset, CIRR. Since ADDE-M strictly relies on attribute labels to construct its memory module, it cannot be directly applied to the CIRR dataset, which lacks attribute annotations. To facilitate the comparison, we adapt ADDE-M by using a pre-trained attribute detector to generate pseudo-labels for CIRR. \textcolor{revisecolor3}{Specifically, to evaluate ADDE-M on CIRR, we utilized Qwen2.5-VL to extract reliable pseudo-labels (achieving 86.8\% Average Precision in our manual inspection of 300 samples). Despite this high quality, a performance gap persists.}
As shown in TABLE~\ref{tab:ADDE-M}, COMBINER significantly outperforms the adapted ADDE-M. This confirms that ADDE-M's reliance on static, explicit attributes leads to its failure in open-domain scenarios, whereas the adaptive semantic disentanglement module designed in COMBINER utilizes a semantic attribute attention mechanism to dynamically disentangle latent prototypes from the current multimodal input. This enables COMBINER to capture fine-grained, unstructured semantics that static memory blocks cannot pre-define (\textit{e.g.}, the foggy atmosphere or rocky texture within the CIRR dataset).}

\begin{table*}[htbp]
  \centering
  \tabcolsep=10pt
    \renewcommand\arraystretch{1.2}
  \caption{Ablation Studies of COMBINER with different components and various settings on FashionIQ, Shoes, and CIRR. Note that AM\# represents different configurations. CIRR-Avg denotes the average of R@$5$ and R$_{subset}$@$1$. The symbol $\Delta$ represents the relative drop in performance for each derivative and is marked with \textcolor{green5}{\textit{the green background}}. \textcolor{defyellowtext}{\textit{The yellow background}} denotes the baseline performance utilized for per column.}
  \vspace{-6pt}
  \resizebox{0.85\linewidth}{!}{
    \begin{tabular}{cc|c|cc|cc|cc|cc|cc}
    \hline\hline
    \multicolumn{1}{c|}{\multirow{3}{*}{AM\#}} & \multicolumn{1}{c|}{\multirow{3}{*}{Var.}} & \multirow{3}{*}{Ablation Methods}  & \multicolumn{6}{c|}{FashionIQ}                & \multicolumn{2}{c|}{\multirow{2}{*}{Shoes}} & \multicolumn{2}{c}{\multirow{2}{*}{CIRR}} \\
\cline{4-9}    \multicolumn{1}{c|}{} & \multicolumn{1}{c|}{} &       & \multicolumn{2}{c|}{Dresses} & \multicolumn{2}{c|}{Shirts} & \multicolumn{2}{c|}{Tops\&Tees} & \multicolumn{2}{c|}{} & \multicolumn{2}{c}{} \\
\cline{4-13}    \multicolumn{1}{c|}{} & \multicolumn{1}{c|}{} &       & Avg.  & $\Delta$  & Avg.  & $\Delta$  & Avg.  & $\Delta$  & Avg.  & $\Delta$  & Avg.  & $\Delta$ \\
    \hline
    \hline
    \rowcolor[rgb]{ .949,  .949,  .949} \multicolumn{13}{c}{\textit{Ablation methods for modules}}\\
    \multicolumn{1}{c|}{1} & \multicolumn{1}{c|}{\multirow{3}{*}{(a)}} &    \multicolumn{1}{l|}{w/o ASD}     & 65.54  & \cellcolor{green2}-2.98  & 70.22  & \cellcolor{green2}-2.52  & 74.63  & \cellcolor{green2}-2.40  & 61.40  & \cellcolor{green2}-2.60  & 79.76  & \cellcolor{green1}-2.16  \\
    \multicolumn{1}{c|}{2} & \multicolumn{1}{c|}{} &  \multicolumn{1}{l|}{w/o UPC}     & 64.77  & \cellcolor{green3}-3.75  & 69.41  & \cellcolor{green3}-3.33  & 74.11  & \cellcolor{green3}-2.92  & 60.62  & \cellcolor{green3}-3.38  & 76.47  & \cellcolor{green4}-5.45  \\
    \multicolumn{1}{c|}{3} & \multicolumn{1}{c|}{} & \multicolumn{1}{l|}{w/o DRM}     & 67.16  & \cellcolor{green1}-1.36  & 71.22  & \cellcolor{green1}-1.52  & 76.11  & \cellcolor{green1}-0.92  & 62.28  & \cellcolor{green1}-1.72  & 79.62  & \cellcolor{green2}-2.30  \\
    \hline
    \rowcolor[rgb]{ .949,  .949,  .949} 
    \multicolumn{13}{c}{\textit{Ablation methods for loss functions}}\\
    \multicolumn{1}{c|}{4} & \multicolumn{1}{c|}{\multirow{6}[2]{*}{(b)}} &        \multicolumn{1}{l|}{w/o cluster-bbc\_t}  & 67.90  & \cellcolor{green2}-0.62  & 71.64  & \cellcolor{green1}-1.10  & 76.19  & \cellcolor{green1}-0.84  & 62.92  & \cellcolor{green2}-1.08  & 81.03  & \cellcolor{green1}-0.89  \\
    \multicolumn{1}{c|}{5} & \multicolumn{1}{c|}{} &        \multicolumn{1}{l|}{w/o cluster-bbc\_c}  & 68.15  & \cellcolor{green1}-0.37  & 71.52  &\cellcolor{green2}-1.22  & 75.62  & \cellcolor{green2}-1.41  & 63.07  & \cellcolor{green1}-0.93  & 80.79  & \cellcolor{green2}-1.13  \\
    \multicolumn{1}{c|}{6} & \multicolumn{1}{c|}{} &        \multicolumn{1}{l|}{w/o cluster-bbc}           & 66.73  & \cellcolor{green3}-1.79  & 70.24  & \cellcolor{green3}-2.50  & 75.11  & \cellcolor{green3}-1.92  & 62.41  & \cellcolor{green3}-1.59  & 80.25  & \cellcolor{green3}-1.67  \\
     \cdashline{3-13}
    \multicolumn{1}{c|}{7} & \multicolumn{1}{c|}{} & \multicolumn{1}{l|}{w/o cluster-kl\_c} & 65.57  & \cellcolor{green2}-2.95  & 69.27  & \cellcolor{green3}-3.47  & 74.32  & \cellcolor{green3}-2.71  & 62.10  & \cellcolor{green1}-1.90  & 78.05  & \cellcolor{green2}-3.87  \\
    \multicolumn{1}{c|}{8} & \multicolumn{1}{c|}{} & \multicolumn{1}{l|}{w/o pool-kl\_p} & 66.26 	&\cellcolor{green1}-2.26 &	70.22 &	\cellcolor{green1}-2.52 &	74.83 &	\cellcolor{green2}-2.20   & 61.04  & \cellcolor{green2}-2.96  & 79.00  & \cellcolor{green1}-2.92  \\
    \multicolumn{1}{c|}{9} & \multicolumn{1}{c|}{} &        \multicolumn{1}{l|}{w/o cluster\&pool-kl}  & 65.02  & \cellcolor{green3}-3.50  & 69.92  & \cellcolor{green2}-2.82  & 75.32  & \cellcolor{green1}-1.71 & 60.76 &	\cellcolor{green3}-3.24   & 77.58 &	\cellcolor{green3}-4.34   \\ 
   \rowcolor{gray!15}
    \multicolumn{3}{l|}{\textbf{COMBINER(Ours)}}        & \textcolor{defblue}{{\textbf{68.52}}}  & \cellcolor{defyellow}0.00  & \textcolor{defblue}{\textbf{72.74}}  & \cellcolor{defyellow}0.00  & \textcolor{defblue}{\textbf{77.03}}  & \cellcolor{defyellow}0.00  & \textcolor{defblue}{\textbf{64.00}}  & \cellcolor{defyellow}0.00  & \textcolor{defblue}{\textbf{81.92}}  & \cellcolor{defyellow}0.00  \\
    \hline\hline
    \end{tabular}%
    }
    \vspace{-15pt}
  \label{tab:ablation}%
\end{table*}%

\subsection{Ablation Study (RQ2)}
\subsubsection{\textcolor{revisecolor}{\textbf{Module Ablation}}}
\textcolor{revisecolor}{To understand the source of these performance gains (RQ2), we conduct a systematic ablation study. We benchmark COMBINER against several ablation variants to dissect the contribution of each component.
}
\noindent
\textit{\textbf{Var.(a) : Ablation methods for modules of COMBINER.}}

\begin{itemize}
\item  \textbf{AM\#1:} We remove the \textit{Adaptive Semantic Disentanglement} module (\textbf{ASD} for short) via
replacing \textit{SAA} with the simple average pooling to obtain the attribute prototype features. 
\item \textbf{AM\#2:} This ablation method removes the \textit{Unified Prototype-based Composition} module (\textbf{UPC} for short) via utilizing simple addition to take the role of features composition. 
\item \textbf{AM\#3:} It evaluates whether \textit{Dual Relations Modeling} module (\textbf{DRM} for short) is effective by removing it from the optimization process.

\end{itemize}

\noindent
\textit{\textbf{Var.(b) : Ablation methods for loss functions of COMBINER.}}

\begin{itemize}
\item \textbf{AM\#4 - AM\#6:} To explore the effect of \textit{cluster-oriented batch-based classification} loss function, we removed $\mathcal{L}_{cls}^t$ and $\mathcal{L}_{cls}^c$ to derive the \textbf{AM\#4} and \textbf{AM\#5}, respectively, and ablated them to emanate the \textbf{AM\#6}.
\item \textbf{AM\#7 - AM\#9:} To investigate the influence of the distribution consistency regularization, we separately removed the cluster-oriented and pool-oriented distribution consistency regularization in Eqn.~(\ref{eq13}) to derive the \textbf{AM\#6} and \textbf{AM\#7} and removed them to obtain the  \textbf{AM\#9}.
\end{itemize}

As shown in TABLE~\ref{tab:ablation}, we draw the following key insights from the ablation experiment results.
\textbf{1)} Removing ASD (\textbf{AM\#1}) causes the performance decrease of COMBINER, showing its important role in improving the performance of COMBINER through disentangling \mbox{attribute prototype} features.
\textbf{2)} \textbf{AM\#2} shows that the variant w/o UPC exhibits the worst performance of all the module variants, which clearly reveals that the UPC module can effectively unify attribute prototypes and alleviate the adverse effects of modal heterogeneity on multimodal query feature composition and metric learning.
\textbf{3)} COMBINER surpasses w/o DRM (\textbf{AM\#3}), indicating that the designed dual relations modeling module can indeed optimize the neighbor relations and improve the model retrieval performance.
\textbf{4)} The performance of \textbf{AM\#4} and \textbf{AM\#5} is inferior to that of COMBINER but superior to that of \textbf{AM\#6}. This indicates that separately pulling multimodal queries and target images closer to the semantic clustering centroids can all optimize neighbor relations. Nevertheless, it is necessary to simultaneously pull both closer to the semantic clustering centroids for better gathering visually similar candidate sets into the same cluster and optimizing their neighbor relations.
\textbf{5)} Ablating either $\mathcal{L}_{kl_c}$ or $\mathcal{L}_{kl_p}$ in Eqn.$($\ref{eq13}$)$ (\textbf{AM\#7}, \textbf{AM\#8}) leads to poor performance. This reveals that the CIR process indeed demands the guidance of target image distribution for multimodal query learning. 
Moreover, removing both cluster-oriented and pool-oriented distribution consistency regularization exhibited the lowest performance, indicating that both are effective in optimizing similarity distribution. This may be due to that they can simultaneously optimize the distributions of the semantic clustering centroids and multimodal queries, resulting in beneficial distribution optimization of neighbor relations.

\begin{table}[h]
\centering
  \tabcolsep=2pt
    \vspace{-8pt}
  \caption{\textcolor{revisecolor}{Performance comparison with the latest SOTA models on the CIRR dataset under the same backbone. The best results under each backbone are \textbf{bolded}.}}
  \vspace{-8pt}
    \renewcommand\arraystretch{1.2}
\label{tab:new_sota_comparison_cirr}
      \resizebox{0.9\linewidth}{!}{
\begin{tabular}{l|cccc|ccc|c}
\hline
    \hline
\multicolumn{1}{c|}{\multirow{2}{*}{Method}} & \multicolumn{4}{c|}{R@$k$}      & \multicolumn{3}{c|}{R$_{subset}$@$k$} & \multirow{2}{*}{Avg} \\
\cline{2-8}         & k=1   & k=5   & k=10  & k=50  & k=1   & k=2   & k=3   &  \\
\hline
\rowcolor{gray!5}
  \multicolumn{9}{c}{\textit{Backbone: CLIP (ResNet50)}}\\
    SADN~\cite{SADN}~\textcolor{gray}{(ACM MM'24)}& 44.27 &	78.10 &87.71 &97.89 &	72.71 &	89.33 &	95.38 	& 75.41  \\

    \multicolumn{1}{l|}{\textbf{COMBINER (Ours)}}  &     {\textbf{45.69}} &	{\textbf{78.79}} &	{\textbf{87.94}} &	{\textbf{97.89}} &	{\textbf{76.33}} &	{\textbf{89.91}} &	{\textbf{96.03}}  & {\textbf{77.56}} \\
        \hline
\rowcolor{gray!5}
  \multicolumn{9}{c}{\textit{Backbone: CLIP (ViT-B)}}\\
    ENCODER~\cite{encoder}~\textcolor{gray}{(AAAI'25)}& 46.10& 77.98& 87.16 &97.64& 76.92& 90.41& 95.95 &77.45\\
\textbf{COMBINER (Ours)} & \textbf{47.21}& 	\textbf{79.11}	& \textbf{88.87} &	\textbf{97.98} 	& \textbf{77.01}& 	\textbf{90.48}	& \textbf{96.24} &  \textbf{78.06}\\
        \hline
\rowcolor{gray!5}
  \multicolumn{9}{c}{\textit{Backbone: CLIP (ViT-L)}}\\
    LIMN+~\cite{limn}~\textcolor{gray}{(TPAMI'24)} &43.33 &	75.41 &	85.81& 	97.21 &	69.28 &	86.43 &	94.26 	&72.35 \\
CoPE~\cite{COPE}\textcolor{gray}{(ACL'25)} & 49.18&	80.65&	89.86&	98.05&	72.34&	88.65&	95.30&	76.49 \\
    DIPNEC~\cite{dipnec}~\textcolor{gray}{(AAAI'25)}& 47.24& 80.20& 89.07 &97.87& 73.97& 89.74& 95.72 &77.09\\
\textbf{COMBINER (Ours)} & \textbf{50.15}&	\textbf{81.83}	&\textbf{89.86}&	\textbf{98.10}	&\textbf{77.76}&	\textbf{91.88}&	\textbf{96.82} & \textbf{79.80} \\
        \hline
\rowcolor{gray!5}
  \multicolumn{9}{c}{\textit{Backbone: CLIP (ViT-H)}}\\
    DQU-CIR~\cite{dqu}~\textcolor{gray}{(SIGIR'24)}& 46.22 &	78.17 &	87.64 &	97.81 	&70.92 &	87.69 &	94.68 &	74.55  \\
\textbf{COMBINER (Ours)} & {\textbf{52.60}} &	{\textbf{82.51}} &	{\textbf{90.12}} &	{\textbf{98.17}} &	{\textbf{81.33}} &	{\textbf{93.06}} &	{\textbf{97.23}}  & {\textbf{81.92}} \\
        \hline
\rowcolor{gray!5}
\multicolumn{9}{c}{\textit{Backbone: BLIP-2 (ViT-G)}}\\
SPRC~\cite{sprc}\textcolor{gray}{(ICLR'24)} & 51.96 &	82.12 &	89.74 &	97.69 &	80.65 &	92.31 &	96.60 	& 81.39 \\
QuRE~\cite{QuRe}\textcolor{gray}{(ICML'25)}  & 52.22&	82.53&	90.31&	98.17&	78.51&	91.28	&96.48&	80.52 \\
\textbf{COMBINER (Ours)} &\textbf{52.63}&	\textbf{82.87}	&\textbf{90.48}	&\textbf{98.41}& \textbf{80.89}	&\textbf{92.96}	&\textbf{97.23 } &\textbf{81.88}\\
\hline
    \hline
\end{tabular}
}
\vspace{-10pt}
\label{tab:new_sota_comparison_cirr-backbone}
\end{table}

\subsubsection{\textcolor{revisecolor}{\textbf{Backbone Ablation}}}
\textcolor{revisecolor}{To demonstrate that the performance gain of COMBINER originates from the innovation in accurately modeling ``attribute-relevant neighbors'', rather than simply relying on a larger-scale backbone network, we conduct a series of rigorous ``same-category backbone'' comparative experiments. As shown in TABLE~\ref{tab:new_sota_comparison_cirr-backbone} and \ref{tab:new_sota_comparison_fiq-backbone}, we adapt COMBINER to five different backbones and compare it with the latest baselines corresponding to those backbones. The results show that when fairly compared with SOTA methods using the same backbone network, COMBINER exhibits consistent and significant superiority. This indicates that the performance advantage of COMBINER is generalizable and decisively stems from the modules we designed, rather than reliance on a specific backbone model.}

\textcolor{revisecolor}{
Meanwhile, for SADN~\cite{SADN}, a method that also utilizes neighbor information, under the identical CLIP~(ResNet-50) backbone network setting, COMBINER surpasses SADN on both FashionIQ (Avg R@10 improvement +$3.47$\%) and CIRR (Avg. +$2.15$\%). This may be because SADN's modeling of neighbor relations primarily relies on global feature similarity, causing it to struggle with distinguishing fine-grained attribute conflicts and be easily misled when faced with distractors with complex backgrounds or highly similar subjects. Whereas COMBINER, through the \textit{Adaptive Semantic Disentanglement (ASD)} module, disentangles fine-grained attribute prototypes, modeling neighbors in the fine-grained attribute space, enabling it to accurately identify and push away those distractors that are visually similar but mismatch in key attributes (such as color, material, object presence). The significant advantage on the most challenging open-domain CIRR dataset strongly confirms the fundamental effect of COMBINER in addressing the core challenge of ``attribute-irrelevant'' interference.}

\begin{table}[h]
  \centering
  \tabcolsep=2pt
        \vspace{-8pt}
    \renewcommand\arraystretch{1.2}
    \caption{\textcolor{revisecolor}{Performance comparison with the latest SOTA models on the FashionIQ dataset under the same backbone. The best results under each backbone are \textbf{bolded}.}}
      \vspace{-8pt}
      \resizebox{0.9\linewidth}{!}{
    \begin{tabular}{l|cc|cc|cc|cc}
    \hline
    \hline    \multicolumn{1}{c|}{\multirow{2}{*}{Method}}     & \multicolumn{2}{c|}{Dresses} & \multicolumn{2}{c|}{Shirts} & \multicolumn{2}{c|}{Tops\&Tees} & \multicolumn{2}{c}{Avg}  \\
\cline{2-9}         & R@$10$  & R@$50$  & R@$10$  & R@$50$  & \multicolumn{1}{c}{R@$10$} & R@$50$  & R@$10$  & R@$50$   \\ \hline 
    \rowcolor{gray!5}
  \multicolumn{9}{c}{\textit{Backbone: CLIP (ResNet50)}}\\
    SADN~\cite{SADN}~\textcolor{gray}{(ACM MM'24)}& 40.01& 65.10& 43.67& 66.05& 48.04& 70.93& 43.91& 67.36  \\
    \multicolumn{1}{l|}{\textbf{COMBINER~(Ours)}} & \textbf{43.33} &	\textbf{68.25} 	&\textbf{46.70} &	\textbf{67.81}  &	\textbf{52.10} &	\textbf{74.76}  &	{\textbf{47.38}} &	{\textbf{70.27}}     \\
        \hline
\rowcolor{gray!5}
  \multicolumn{9}{c}{\textit{Backbone: CLIP (ViT-B)}}\\
    ENCODER~\cite{encoder}~\textcolor{gray}{(AAAI'25)}& 51.51& 76.95& 54.86& 74.93& 62.01& 80.88& 56.13& 77.59  \\
    \multicolumn{1}{l|}{\textbf{COMBINER~(Ours)}} & \textbf{52.90} &	\textbf{76.70} 	&\textbf{55.94} &	\textbf{75.66}  &	\textbf{62.93} &	\textbf{82.76}  &	{\textbf{57.26}} &	{\textbf{78.37}}     \\
        \hline
\rowcolor{gray!5}
  \multicolumn{9}{c}{\textit{Backbone: CLIP (ViT-L)}}\\
  LIMN+~\cite{limn}~\textcolor{gray}{(TPAMI'24)} &52.11 &75.21& 57.51& 77.92& 62.67& 82.66&57.43 &	78.60  \\
CoPE~\cite{COPE}\textcolor{gray}{(ACL'25)} & 39.85	&66.98	&45.03&	66.81&	48.61&	72.01	&44.5	&68.6   \\
DIPNEC~\cite{dipnec}\textcolor{gray}{(AAAI'25)} & 46.90 &71.29 &56.92& 77.77& 58.18& 80.88& 54.00& 76.64   \\
\multicolumn{1}{l|}{\textbf{COMBINER~(Ours)}} & \textbf{53.45} 	& \textbf{77.29} 	& \textbf{59.42} & 	\textbf{79.29} & 	\textbf{64.46} & 	\textbf{84.04} & 	\textbf{59.11} & 	\textbf{80.21}      \\
        \hline
\rowcolor{gray!5}
  \multicolumn{9}{c}{\textit{Backbone: CLIP (ViT-H)}}\\
    DQU-CIR~\cite{dqu}~\textcolor{gray}{(SIGIR'24)}& {57.63} &	{78.56} 	&{62.14} &	{80.38} &	{66.15} &	{85.73} &	{61.97} &	{81.56} \\
    \multicolumn{1}{l|}{\textbf{COMBINER~(Ours)}} & {\textbf{57.96}} &	{\textbf{79.08}} &	{\textbf{63.64}} &	{\textbf{81.84}} &	{\textbf{68.18}} &	{\textbf{85.87}} &	{\textbf{63.26}} &	{\textbf{82.27}}     \\
        \hline
\rowcolor{gray!5}
\multicolumn{9}{c}{\textit{Backbone: BLIP-2 (ViT-G)}}\\
SPRC~\cite{sprc}\textcolor{gray}{(ICLR'24)}& 49.18 &	72.43 &	55.64 &	73.89 	&59.35 &	78.58 &	54.72 &	74.97 \\
QuRE~\cite{QuRe}\textcolor{gray}{(ICML'25)}  & 46.80	&69.81	&53.53	&72.87&	57.47&	77.77&	52.6	&73.48   \\
    \multicolumn{1}{l|}{\textbf{COMBINER~(Ours)}} & \textbf{53.20} 	&\textbf{76.70} 	&\textbf{63.10} &	\textbf{81.45} &	\textbf{66.50} &	\textbf{84.40} &	\textbf{60.93} 	&\textbf{80.85}    \\
    \hline    \hline
    \end{tabular}
}
\vspace{-4pt}
  \label{tab:new_sota_comparison_fiq-backbone}%
\end{table}%

\subsubsection{\textcolor{revisecolor}{\textbf{Loss Ablation}}}
\textcolor{revisecolor}{To demonstrate that our proposed DRM module is not merely a simple Deep Metric Learning (DML) enhancement, we conducted a critical experiment, replacing our core losses with two representative DML losses:}
\begin{itemize}[leftmargin=14pt]
      \item \textcolor{revisecolor}{\textbf{COMBINER w/ HIST:} We replace $\mathcal{L}_{cls}$, $\mathcal{L}_{kl_p}$, and $\mathcal{L}_{kl_c}$ in COMBINER with $\mathcal{L}_{D}$ and $\mathcal{L}_{CE}$ from HIST~\cite{R-3}.}
    \item \textcolor{revisecolor}{\textbf{COMBINER w/ AFNE:} We replace $\mathcal{L}_{cls}$, $\mathcal{L}_{kl_p}$, and $\mathcal{L}_{kl_c}$ in COMBINER with $\mathcal{L}^{AFNE}$ from AFNE~\cite{R-4}.}
    \end{itemize}
\textcolor{revisecolor}{
TABLE~\ref{tab:dml} shows the performance comparison between the full COMBINER model and its variants based on DML losses on the FashionIQ and CIRR datasets. We can clearly observe that, compared to the full COMBINER, both DML variants exhibit a significant performance degradation. This is particularly evident on FashionIQ-Avg (R@10), where COMBINER w/ HIST shows a drop exceeding 7\%.
Regarding COMBINER w/ AFNE, AFNE is designed to handle False Negatives through adaptive weighting. However, the `attribute-irrelevant samples' we target (such as the `carpet' in Fig.~\ref{fig:intro}(c)) are genuine negative samples; they are semantically incorrect, merely visually deceptive. The mechanism of AFNE erroneously treats these samples as `potential positive samples,' which consequently leads to incorrect optimization.
This experiment demonstrates that generic DML losses~\cite{R-3, R-4} cannot address the unique challenge of `attribute-irrelevant negative samples' in CIR. In contrast, our proposed DRM module is specifically designed to resolve this particular issue, and is therefore both necessary and irreplaceable.
}

\begin{table}[ht]
      \centering
      \renewcommand\arraystretch{1.2}
              \vspace{-7pt}
      \caption{\textcolor{revisecolor}{Performance comparison between COMBINER and its variants based on DML losses on the FashionIQ and CIRR datasets. CIRR-Avg denotes the average of R@5 and R$_{subset}$@1. The best results are shown in bold.}}
        \vspace{-7pt}
      \tabcolsep=4pt
      \resizebox{0.9\linewidth}{!}{
        \begin{tabular}{l|c|c|c}
          \hline
          \hline
          \multicolumn{1}{c|}{Method} & FashionIQ-Avg (R@10) & FashionIQ-Avg (R@50) & CIRR-Avg \\
          \hline
          COMBINER w/ HIST & 56.13 & 76.22 & 74.46 \\
          COMBINER w/ AFNE & 57.98 & 77.86 & 76.15 \\
          \rowcolor{gray!15}
          \textbf{COMBINER (Ours)} & \textbf{63.26} & \textbf{82.27} & \textbf{81.92} \\
          \hline
          \hline
        \end{tabular}
      }
      \vspace{-8pt}
      \label{tab:dml}
    \end{table}

\subsubsection{\textcolor{revisecolor}{\textbf{Clustering Algorithm Ablation}}}
\textcolor{revisecolor}{To further investigate the influence of the clustering algorithm utilized for constructing semantic clusters in the DRM module, we adapted K-Means and three other mainstream clustering algorithms to COMBINER, applying the same number of clusters, $H$, for comparison. Our variant designs are as follows:}
\begin{itemize}[leftmargin=8pt]
    \item \textcolor{revisecolor}{\textbf{COMBINER w/ DBSCAN:} utilizes DBSCAN, which is a classic density-based clustering algorithm, where clusters are partitioned based on the connectivity of samples in high-density regions.}
\item \textcolor{revisecolor}{\textbf{COMBINER w/ Agg:} utilizes Hierarchical Clustering, specifically, a ``bottom-up'' hierarchical aggregation strategy.}
\item \textcolor{revisecolor}{\textbf{COMBINER w/ Spec:} utilizes Spectral Clustering, which is based on graph theory, first projecting the data into a lower-dimensional space before performing clustering.}
\end{itemize}

\begin{table}[ht]
      \centering
\renewcommand\arraystretch{1.2}
        \vspace{-7pt}
\caption{\textcolor{revisecolor}{Performance comparison between COMBINER and its variants using different clustering algorithms on the FashionIQ and CIRR datasets. CIRR-Avg denotes the average of R@5 and R$_{subset}$@1. The best results are shown in bold.}}
      \tabcolsep=4pt
        \vspace{-6pt}
      \resizebox{0.9\linewidth}{!}{
        \begin{tabular}{l|c|c|c}
          \hline
          \hline
          \multicolumn{1}{c|}{Method} & FashionIQ-Avg (R@10) & FashionIQ-Avg (R@50) & CIRR-Avg \\
          \hline
          COMBINER w/ DBSCAN & 61.05 & 80.19 & 79.43  \\
          COMBINER w/ Agg &62.93 & \textbf{82.31} & 81.55  \\
          COMBINER w/ Spec & 61.32 & 80.44 & 79.80 \\
          \rowcolor{gray!15}
          \textbf{COMBINER (Ours)} & \textbf{63.26} & 82.27 & \textbf{81.92} \\
          \hline
          \hline
        \end{tabular}
      }
      \vspace{-8pt}
      \label{tab:clustering}
    \end{table}
\textcolor{revisecolor}{
TABLE~\ref{tab:clustering} illustrates the impact of different clustering algorithms on model performance. We found that COMBINER w/ Agg (Hierarchical Clustering) performs comparably to the full model (K-Means). However, both COMBINER w/ DBSCAN and COMBINER w/ Spec (Spectral Clustering) exhibited a noticeable performance degradation.
}
\textcolor{revisecolor}{
We attribute this performance degradation to the mismatch between the core mechanisms of these algorithms and the requirements of our task. For DBSCAN, which is a density-based algorithm, it faces the challenge of the ``curse of dimensionality'' when dealing with high-dimensional and sparse feature spaces (such as CLIP embeddings). This makes DBSCAN highly prone to mistakenly classifying many meaningful samples as ``noise,'' thereby compromising the integrity of proximity relationships.}
\textcolor{revisecolor}{
Conversely, for Spectral Clustering, which relies on graph construction, constructing a graph matrix that accurately reflects semantic properties rather than visual appearance similarity is extremely difficult in high-dimensional spaces. The clusters found by Spectral Clustering may correspond to arbitrary shapes on the feature manifold, which is inconsistent with our desired attribute clusters that span different visual categories (\textit{e.g.}, ``all black items'').
}

\subsubsection{\textcolor{revisecolor}{\textbf{Alignment Strategy Ablation}}}
\textcolor{revisecolor}{The mutual-learning strategy within the DRM module depends on KL divergence (Kullback-Leibler Divergence) to align the distributions of the composed features (Query) and the target features (Target). To evaluate the rationale of this selection, we replace it with other mainstream distribution alignment strategies for comparison. Our variant designs are as follows:}
\begin{itemize}
    \item  \textcolor{revisecolor}{\textbf{COMBINER w/ JSD}: We replace KL divergence with JS divergence (Jensen-Shannon Divergence), which is a symmetric and smoothed version of KL divergence.}
    \item \textcolor{revisecolor}{\textbf{COMBINER w/ OT}: We replace KL divergence with Wasserstein-1 distance, also known as Optimal Transport (OT) distance, which provides meaningful gradients when measuring non-overlapping distributions.}
\end{itemize}

\begin{table}[ht]
      \centering
      \vspace{-8pt}
\renewcommand\arraystretch{1.2}
\caption{\textcolor{revisecolor}{Performance comparison between COMBINER and its variants using different distribution alignment strategies on the FashionIQ and CIRR datasets. CIRR-Avg denotes the average of R@5 and R$_{subset}$@1. The best results are in \textbf{bold}.}}
  \vspace{-7pt}
      \tabcolsep=4pt
      \resizebox{0.98\linewidth}{!}{
        \begin{tabular}{l|c|c|c}
          \hline
          \hline
          \multicolumn{1}{c|}{Method} & FashionIQ-Avg (R@10) & FashionIQ-Avg (R@50) & CIRR-Avg \\
          \hline
          COMBINER w/ JSD & 62.48 & 81.69 & 81.21  \\
          COMBINER w/ OT & 62.33 & 81.50 & 80.95 \\
          \rowcolor{gray!15}
          \textbf{COMBINER (Ours)} & \textbf{63.26} & \textbf{82.27} & \textbf{81.92} \\
          \hline
          \hline
        \end{tabular}
      }
      \vspace{-8pt}
      \label{tab:alignment}
    \end{table}
\textcolor{revisecolor}{
As shown in TABLE~\ref{tab:alignment}, we observe that the default KL divergence (Ours) significantly outperforms JS divergence and OT distance on all metrics. This result strongly supports our design rationale. Our mutual-learning strategy is essentially a form of Knowledge Distillation, aiming to achieve Uni-directional Supervision: specifically, using the distribution of the target features (as Teacher) to guide the distribution of the composed features (as Student). It strictly penalizes cases where the Student distribution predicts regions not covered by the Teacher distribution, thereby forcing the distribution of composed features to fit the distribution of target features.
}
\textcolor{revisecolor}{
Conversely, JS divergence brings the two distributions together symmetrically. This ``bi-directional'' characteristic instead weakens the specific guiding role of the target feature distribution as the ``Teacher,'' leading to a performance degradation. Additionally, for OT, although it is robust in handling non-overlapping distributions, it primarily focuses on minimizing the ``transportation cost'' between distributions. In this task, we are more concerned with the precise alignment of the two distributions in terms of probability density, rather than the cost of ``moving.'' Therefore, the entropy-based KL divergence performs better.
}

\subsection{Sensitivity Analysis (RQ3)}
\begin{figure}[h]
    \centering
     \vspace{-5pt}
	\includegraphics[width=0.9\linewidth]{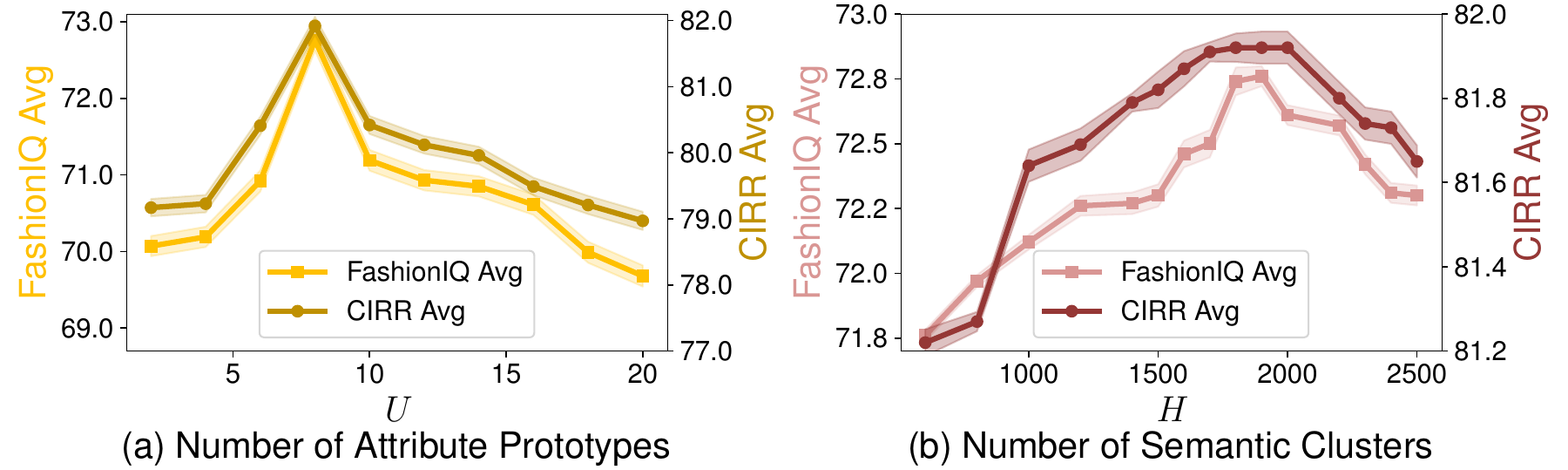}
\vspace{-6pt}
	\caption{Influence of (a) Attribute Prototype Number $U$ and (b) Semantic Cluster Number $H$ on FashionIQ and CIRR datasets.}
\vspace{-4pt}
	\label{fig:num_cluster}
\end{figure}

In this section, to exhibit more insight into the sensitivities of COMBINER to hyper-parameters, we exhibit the performance comparison with different trade-off hyper-parameters settings and numbers of attribute prototypes and semantic clusters in Fig.~\ref{fig:loss_sen} and Fig.~\ref{fig:num_cluster}, respectively.

\subsubsection{Sensitivity to Trade-off Hyper-parameters}
\begin{figure*}[h]
    \centering
	\includegraphics[width=0.9\linewidth]{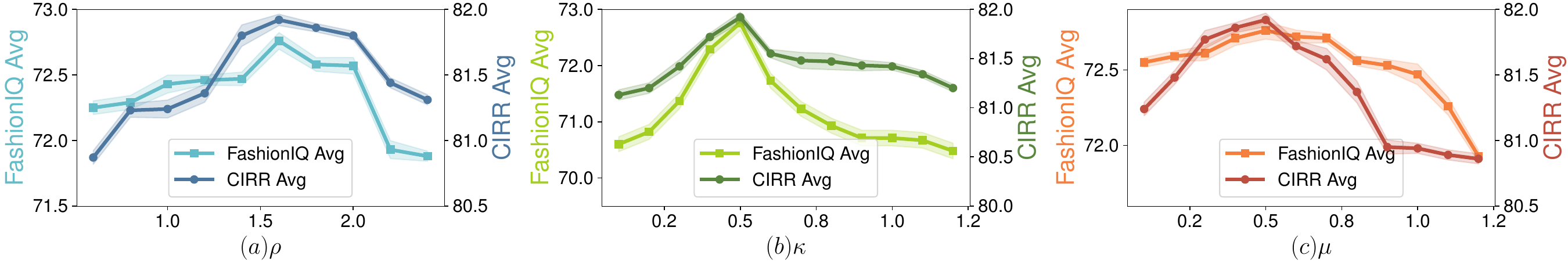}
    \vspace{-9pt}
	\caption{Influence of hyper-parameters $\rho$, $\kappa$, and $\mu$ on FashionIQ and CIRR datasets.}
    \vspace{-15pt}
	\label{fig:loss_sen}
\end{figure*}

This part aims to investigate the sensitivities of COMBINER to the (a) the cluster-oriented batch-based classification loss hyper-parameter $\rho$, (b) the cluster-oriented distribution consistency regularization loss hyper-parameter $\kappa$, and (c) the pool-oriented distribution consistency regularization hyper-parameter $\mu$. 
Drawing on the outcomes presented in Fig.~\ref{fig:loss_sen}, we can highlight the following principal findings.
\textbf{1)} For $\rho$, performance rises slowly on \mbox{FashionIQ} while it rises quickly on CIRR, both reaching high levels and then gradually declining. This indicates that the cluster-oriented batch-based classification loss is effective in promoting features with similar attributes, but too large $\rho$ may lead to incorrectly advancing visually similar samples and over-pushing away the corresponding target images.
\textbf{2)} For $\kappa$, the performance on the two datasets first rises gradually and then declines with increasing $\kappa$. This suggests that the pool-oriented distribution consistency regularization indeed aids in learning the similarity distribution of multimodal composed features. However, an excessively large $\kappa$ may cause the distribution to overly tend to the target image distribution within the batch, leading to a decline in performance.
\textbf{3)} For $\mu$, performance on FashionIQ is consistently high, dropping slightly at a higher value of $\mu$. On CIRR, the performance first rises and then decreases as $\mu$ increases. 
This exhibits that it does converge the distributions of the composed feature and the target feature, but too large $\mu$ may lead to excessive false convergence.

\subsubsection{Sensitivity to Attribute Prototype and Semantic Cluster Number}
Furthermore, to study the sensitivity of COMBINER to the attribute prototype number $U$ and the number of semantic clusters $H$, we present a performance comparison of COMBINER on FashionIQ and CIRR datasets with different numbers of attribute prototypes and semantic clusters in Fig.~\ref{fig:num_cluster}. From this figure, we can reveal the following key insights.

For attribute prototype number \textit{$U$} of attribute-level features in COMBINER, the model's performance follows a consistent pattern as \textit{$U$} increases. After a certain number of attribute prototypes are caught, COMBINER slightly decreases and then levels off. Note that the performance declines when the number of attribute prototypes is too high. This is reasonable that excessive prototypes in attribute-level features can lead to disruptions in subsequent neighbor relations modeling. Furthermore, on both the FashionIQ and CIRR datasets, COMBINER achieves its optimal performance at $U=8$. This demonstrates that the attribute prototype number in COMBINER is robust to variations across datasets and no significant adjustment is required to encompass the prototype attributes of different datasets at this setting. \textcolor{revisecolor}{As shown in Fig.~\ref{fig:num_cluster} (a), when $U=8$, the performance of COMBINER on both the FashionIQ and CIRR datasets reaches its optimum. This indicates that a capacity of $U=8$ is sufficient to handle the primary attribute modifications in the CIR task, regardless of whether it is in the fashion domain or the open domain. Furthermore, we state in the implementation details section (Section~\ref{exp:experimental_settings}) that COMBINER uniformly uses $U=8$ across all datasets. Therefore, $U$ is a parameter related to the task rather than being dataset-bound, possessing good generalization capability.}

For the number of semantic clusters $H$, COMBINER's performance on the two datasets first increases and then decreases with minor fluctuations, as $H$ increases. This is reasonable since an insufficient number of cluster centers may lead to inadequate prototype learning within the datasets, thereby degrading retrieval performance.
Moreover, the difference between their maximum and minimum values does not exceed $1.0$. These findings imply that COMBINER's retrieval performance remains robust to changes of $H$ within the tested range. 
\textcolor{revisecolor}{This strongly supports our design choices. Specifically, we state in the implementation details section (Section~\ref{exp:experimental_settings}) that COMBINER uniformly uses $H=1900$ across all datasets and achieves SOTA performance on all datasets. This fact strongly demonstrates that $H$ is not a sensitive parameter requiring fine-tuning for specific datasets, possessing strong generalizability across different datasets.}
The stability may indicate that COMBINER effectively captures attribute-based neighbor relations across various cluster sizes.

\subsection{Inference Efficiency Analysis (RQ4)}
\textcolor{revisecolor}{Beyond retrieval accuracy, real-world applications also demand high efficiency. We now analyze the inference speed of COMBINER.}
We present in TABLE~\ref{tab:inference} a comparison of the inference efficiency between the proposed COMBINER and the representative Composed Image Retrieval (CIR) model DQU-CIR~\cite{dqu}. Specifically, the table reports the external models utilized, inference time, and the corresponding retrieval performance, which were all run on a single NVIDIA A40 GPU.
Based on these outcomes, we arrive at the following conclusions.
Although DQU-CIR leverages additional external models, this results in a significant decline in inference efficiency. In contrast, COMBINER achieves a substantial improvement in retrieval performance while maintaining high inference efficiency. \textcolor{revisecolor2}{More precisely, COMBINER reduces the per-sample inference time by $98.78$\% largely because our core training modules, such as the K-Means clustering in DRM, are stripped away during inference, and the remaining ASD and UPC modules add negligible FLOPs to the backbone. This allows COMBINER to obtain retrieval accuracy on CIRR-Avg with $9.89$\% improvement.}
These findings demonstrate that COMBINER does not require external models to enhance the understanding of multimodal queries. Instead, through solely learning learning the designed attribute-based neighbor relations, COMBINER's ability to discriminate among visually similar samples can be enhanced (\textit{i.e.,} COMBINER can push apart visually similar samples with differing attributes while pulling together samples that share the same attributes), thereby effectively improving retrieval performance.
\begin{table}[t]
  \centering
      \renewcommand\arraystretch{1.2}
  \caption{Comparison of inference efficiency. The better results are in bold. Infer denotes the inference time for each sample. FIQ-R$10$-Avg is the mean R@$10$ value on FashionIQ and CIRR-Avg represents the ((R@$5$+R$_{subset}$@$1$)/$2$) on CIRR.}
    \resizebox{0.9\linewidth}{!}{
    \begin{tabular}{c|c|c|c|c}
    \hline
    \hline
    Methods & External Models & Infer$\downarrow$ & FIQ-R$10$-Avg$\uparrow$ & CIRR-Avg$\uparrow$ \\

    \hline
    DQU-CIR & BLIP-2+Gemini-pro-v1 & 2.045s & 61.97 & 74.55 \\
       \rowcolor{gray!15}
    \textbf{COMBINER(Ours)} & \textcolor{defblue}{\textbf{\usym{1F5F6}}} & \textcolor{defblue}{\textbf{0.025s}} & \textcolor{defblue}{\textbf{63.26}} & \textcolor{defblue}{\textbf{81.92}} \\
    \hline
    \hline
    \end{tabular}%
    }
  \label{tab:inference}%
\end{table}%

\subsection{Qualitative Analysis (RQ5)}
\textcolor{revisecolor}{In order to provide an intuitive understanding of our model's behavior, we present qualitative case studies and visualizations in the following.}
\subsubsection{Case Study}
\begin{figure}[ht!]
    \centering
    \vspace{-8pt}
	\includegraphics[width=0.92\linewidth]{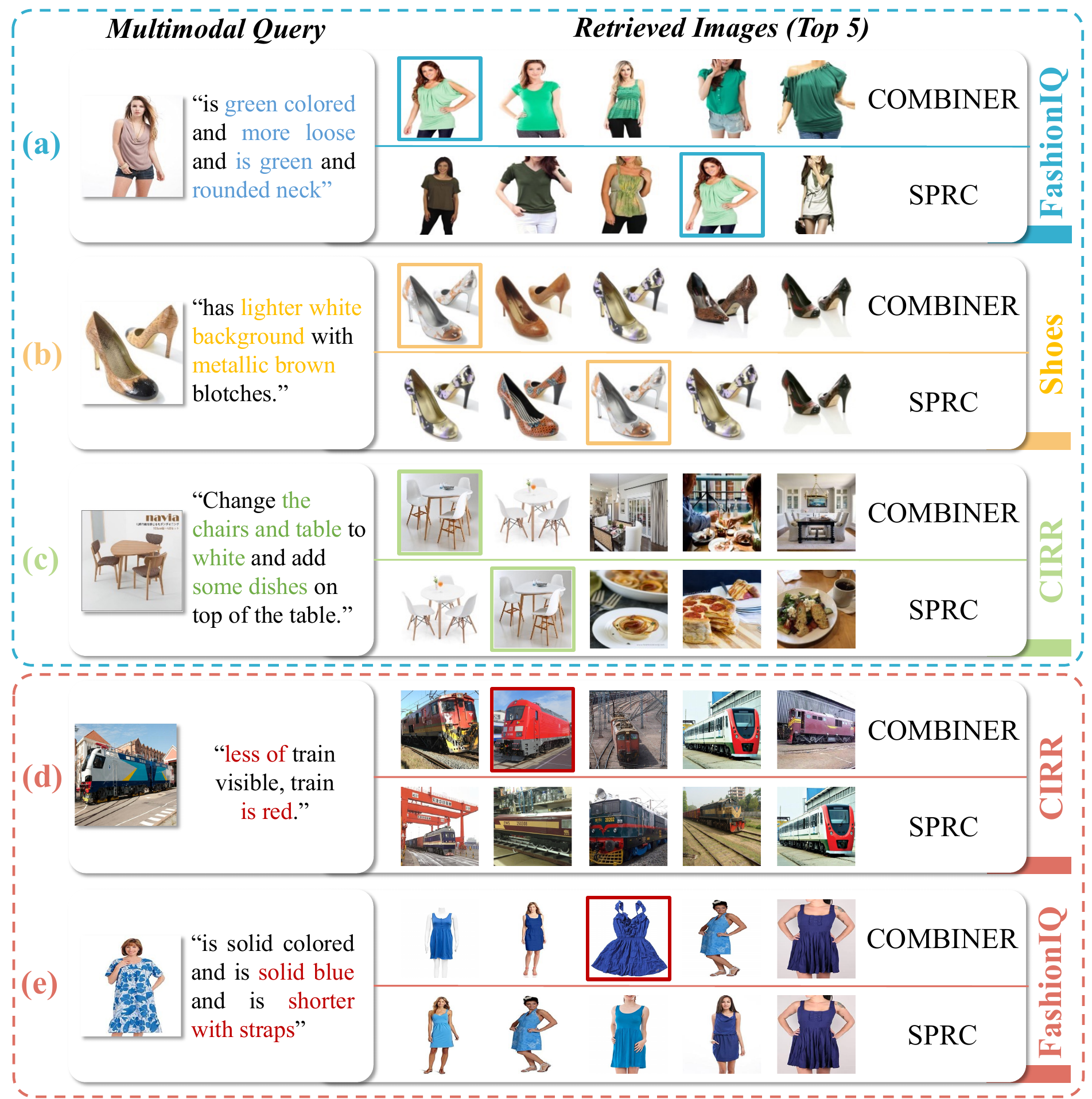}
    \vspace{-6pt}
	\caption{Case study on (a) FashionIQ, (b) Shoes, (c) CIRR, \textcolor{revisecolor}{and failure cases on (d) CIRR and (e) FashionIQ.}}
\vspace{-8pt}
	\label{fig:case_study}
\end{figure}

\textcolor{revisecolor3}{As illustrated in Fig.~\ref{fig:case_study}, we provide a qualitative comparison with SPRC, which serves as a highly representative open-source state-of-the-art method. It can be observed that while the open-source baseline frequently struggles with the ``visually similar but attribute-unrelated'' dilemma, COMBINER accurately identifies the precise target images. This directly visualizes the superiority of our explicitly modeled attribute-based neighbor relations in resisting visually deceptive negative samples.}
Specifically, as shown in Fig.~\ref{fig:case_study}(a), COMBINER successfully retrieves the optimal image by leveraging learned attribute-based neighbor relations to capture text modifications (e.g., ``green colored'', ``more loose'') while preserving reference details (``garment folds''). In contrast, SPRC misses the more loose'' requirement and alters the original style, likely because its sentence-level prompt mechanism struggles to capture fine-grained attributes. Furthermore, in Fig.~\ref{fig:case_study}(b) and (c), COMBINER effectively integrates multiple attributes simultaneously (e.g., ``lighter white'' and ``metallic brown'', or ``white'' and ``dishes''). Conversely, SPRC's top-ranked results only partially satisfy these queries by capturing single attributes independently, demonstrating its inability to comprehensively model complex multi-modal queries.
\textcolor{revisecolor}{In addition to the successful cases discussed above, to analyze the model's limitations more comprehensively, we also present a representative failure case in Fig.~\ref{fig:case_study}(d). In this case, COMBINER fails to recall the specified ``Ground-Truth'' target image among the Top-K results. However, upon careful inspection, we discover that the Top-K results retrieved by COMBINER (for example, the first few images shown in the figure) fully satisfy the attribute requirements of the modification text in a semantic sense (\textit{e.g.}, correctly ``removing the train's original blue color'' and ``replacing it with red''). This phenomenon reveals the issue of ``false-negatives'' existing in the dataset's annotations: specifically, the retrieved images are semantically correct matches but were not labeled as the target image. Therefore, although our model correctly comprehended the query intention, its R@k metrics were penalized as a result of such incomplete annotations.}
\textcolor{revisecolor3}{As illustrated in the Fig.~\ref{fig:case_study}(e), COMBINER occasionally struggles with highly coupled attributes, where a visually dominant feature (e.g., ``solid blue'') overshadows fine-grained structural details (e.g., ``shorter with straps''). This reveals a capacity boundary in our ASD module for disentangling perfectly balanced prototypes under extreme feature dominance, guiding our future optimization efforts.}
These observations underscore the necessity for COMBINER to maintain attribute prototype features when modeling neighbor relations. The composition of attribute-level information with neighbor relations yields superior retrieval results and enhances COMBINER's accuracy in the CIR task.

\subsubsection{Similarity Matrix Visualization}
\label{appendix:matrix}
\begin{figure}[ht]
    \centering
	\includegraphics[width=0.9\linewidth]{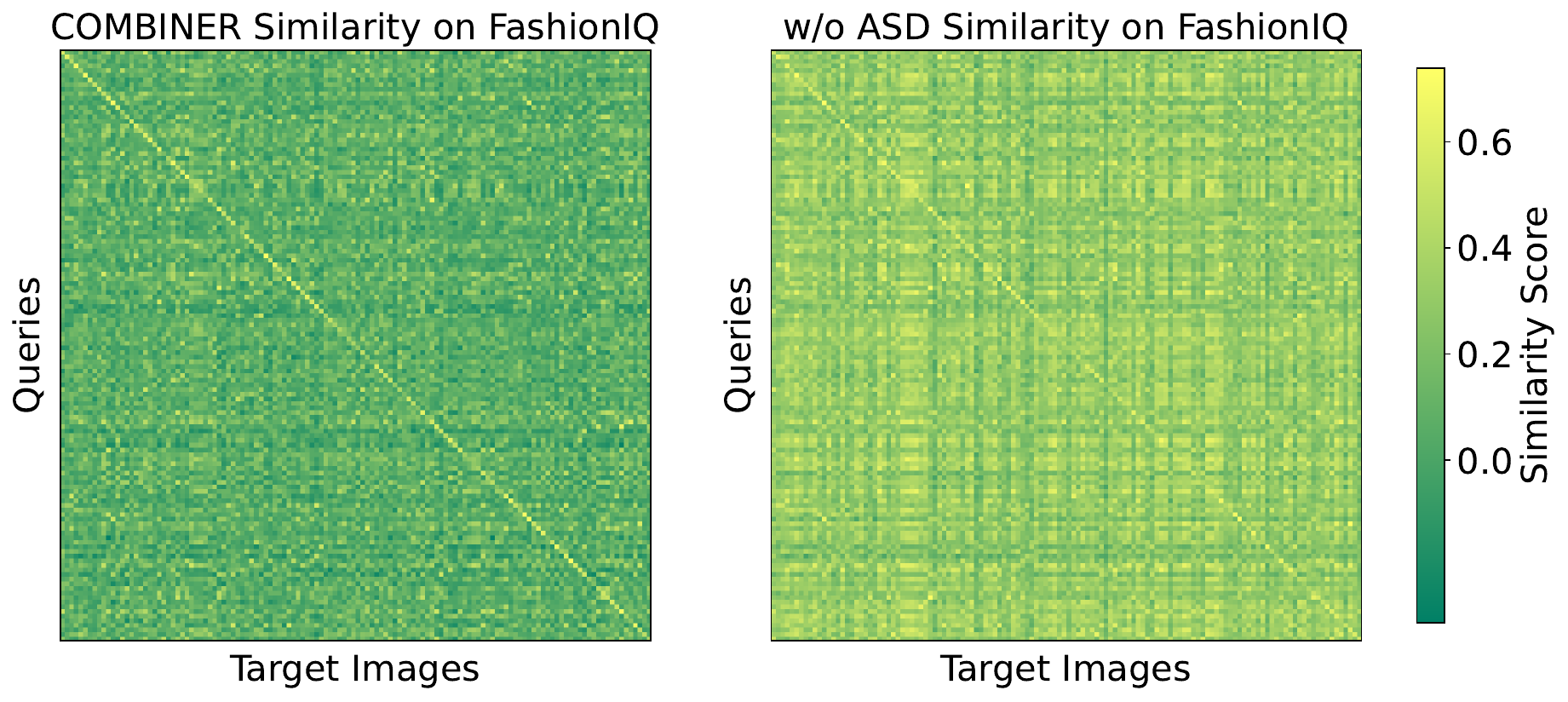}
    \vspace{-8pt}
	\caption{Similarity Matrix Visualization on FashionIQ.}
    \vspace{-6pt}
	\label{fig:sim_mat}
\end{figure}

\noindent The similarity matrices shown in Fig.~\ref{fig:sim_mat} offer a visual comparison between COMBINER and its variant w/o ASD, revealing distinct patterns in their relationships between multimodal queries and target images.
COMBINER's matrix shows a more pronounced contrast between predominantly darker tones and brighter main diagonals. This pattern suggests that COMBINER achieves greater differentiation, maintaining high similarity scores between each multimodal query and its corresponding target image, while effectively reducing similarities with other candidates.
Conversely, the variant w/o ASD displays a more uniform distribution of lighter green hues across the matrix. This indicates that without ASD, the model retains relatively high similarity scores between queries and a wider range of candidate images, potentially leading to less precise matching.

\subsubsection{Attention Visualization}

\begin{figure}[h]
    \centering
        \vspace{-10pt}
	\includegraphics[width=0.9\linewidth]{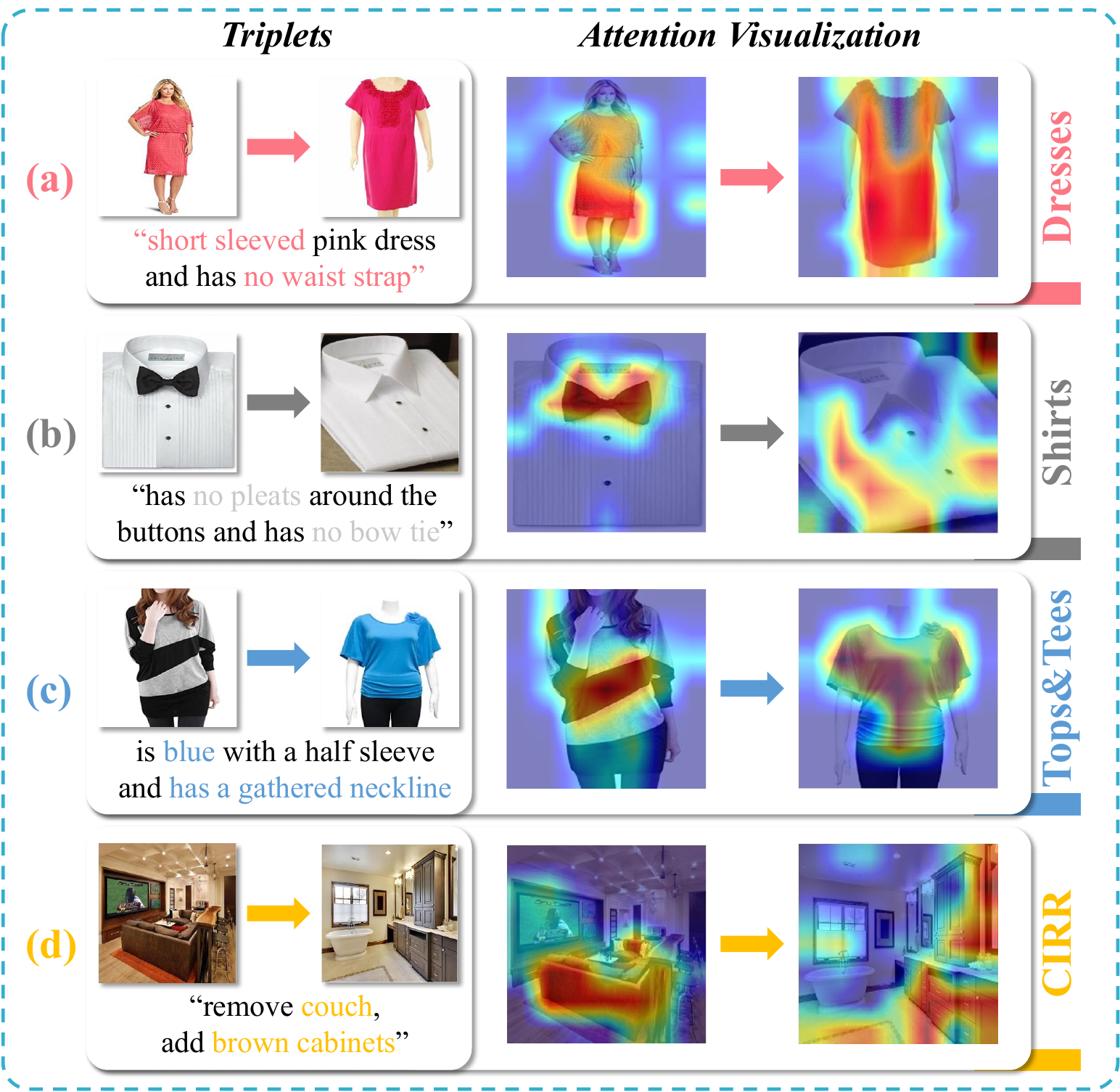}
    \vspace{-6pt}
	\caption{Attention Visualizations on (a) Dresses, (b) Shirts, (c) Tops\&Tees, and (d) CIRR datasets.}
    \vspace{-8pt}
	\label{fig:attention}
\end{figure}

Fig.~\ref{fig:attention} illustrates the attention distribution of attribute-level features on the reference and target images in \mbox{FashionIQ} and CIRR triplet samples, as visualized via the GradCAM method~\cite{grad-cam}. 
In example (a), the model attends to ``short sleeved'' in the target image and ``no waist strap'' in the reference image, demonstrating COMBINER's ability to faithfully interpret and incorporate the semantic information from the modification text.
In sample (b), the attribute‐level embeddings concentrate on the ``bow tie'' region of the reference image and the ``pleats'' region of the retrieved image, exactly matching the semantics provided by the modification text.
Furthermore, sample (c) demonstrates that COMBINER is capable of identifying the area corresponding to the global modification information in the modification text, as evidenced by the correct marking of ``blue'' and ``gathered neckline'' in the target and reference image. As for sample (d), the attribute‐level features correctly pinpoint the ``couches'' region in the reference image and the ``brown cabinets'' region in the retrieved image, exactly reflecting the modification text's intent. These observations demonstrate that COMBINER is capable of effectively focusing on the modified region corresponding to the modification text in the image.

\subsubsection{Visualization of Semantic Cluster Neighbors}
\begin{figure}[h]
    \centering
	\includegraphics[width=0.93\linewidth]{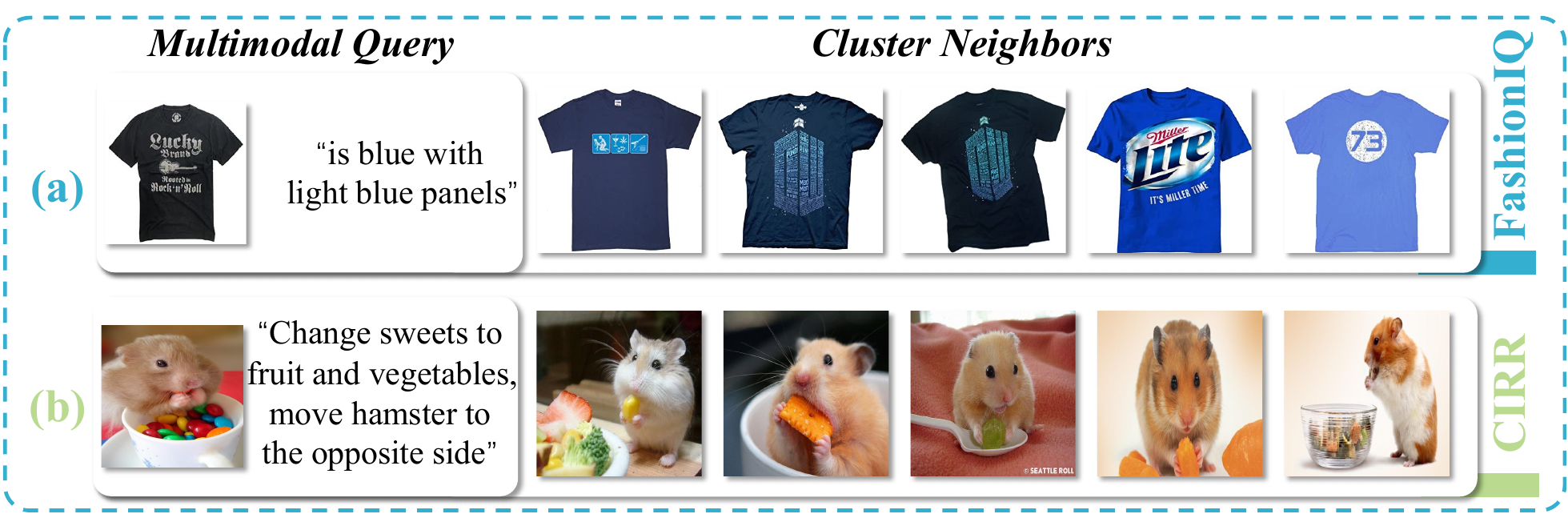}
    \vspace{-6pt}
	\caption{Visualization of Semantic Cluster Neighbors on (a) FashionIQ and (b) CIRR datasets.}
    \vspace{-2pt}
	\label{fig:cluster}
\end{figure}

\textcolor{revisecolor2}{
To further validate the effectiveness of the Dual Relations Modeling (DRM) module in capturing semantically consistent neighbor relations, we conducted a K-means clustering analysis on the target features extracted by the trained COMBINER. In Fig.~\ref{fig:cluster}, we visualize the neighbor samples within the semantic clusters corresponding to specific multimodal queries. As shown in row (a) (FashionIQ dataset), although the reference image is a black t-shirt, the modification text explicitly requests ``is blue with light blue panels''. The visualized cluster neighbors consistently exhibit this specific ``blue'' attribute, accurately capturing the attribute modification. Similarly, in row (b) (CIRR dataset), the query requests to ``Change sweets to fruit and vegetables''. Despite the reference image showing a hamster with sweets, the neighbors within the cluster consistently depict hamsters interacting with fruits or vegetables. This qualitative result demonstrates that COMBINER successfully groups samples that are ``visually similar and attribute-related'' while effectively filtering out attribute-unrelated noise. This validates the efficacy of the DRM module in optimizing the semantic structure of the metric space.
}

%% file: 5_con.tex
\section{Conclusion}
In this work, to mitigate the detrimental effects of visually similar but attribute-unrelated samples on metric learning, which were widespread in CIR tasks, we proposed a new unified representation of cross-modal features based on attribute prototypes that optimized multimodal feature fusion and metric learning. Building on this foundation, we introduced COMBINER, which comprised three main modules. Firstly, we designed an Adaptive Semantic Disentanglement module that was capable of disentangling attribute features based on multimodal primitive features. Secondly, we proposed a Unified Prototype-based Composition module that could construct cross-modal unified prototypes (CUP) and facilitate multimodal feature composition. Finally, we introduced a Dual Relations Modeling module that uncovered pairwise and neighbor relations based on attribute similarity. Our thorough evaluation of three standard benchmarks confirms that COMBINER consistently outperforms existing methods. In the future, we plan to further extend our approach for intelligent customer service robots.